\documentclass[10pt,twocolumn,letterpaper]{article}

\usepackage[pagenumbers]{cvpr} 

\usepackage{graphicx}
\usepackage{amsmath}
\usepackage{amsthm}
\usepackage{amssymb}
\usepackage{booktabs}
\usepackage{enumerate}
\usepackage{multicol}
\usepackage{mathtools}
\usepackage{multirow}
\usepackage{amssymb}
\usepackage{pifont}
\newcommand{\cmark}{\ding{51}}%
\newcommand{\xmark}{\ding{55}}%
\usepackage{overpic}

\usepackage[pagebackref,breaklinks,colorlinks]{hyperref}

\usepackage[capitalize]{cleveref}
\crefname{section}{Sec.}{Secs.}
\Crefname{section}{Section}{Sections}
\Crefname{table}{Table}{Tables}
\crefname{table}{Tab.}{Tabs.}

\newcommand{\psd}{Z}
\newcommand{\point}{X}
\newcommand{\hompoint}{\bar X}
\newcommand{\obs}{\tilde x}
\newcommand{\repr}{x}
\newcommand{\homrepr}{\bar x}
\newcommand{\onevar}{\alpha}

\newcommand{\cat}{;} 

\newcommand{\tr}{\text{tr}}

\newcommand{\psdset}[1]{\mathbb{S}^{#1}_{+}}
\newcommand{\SM}{\text{BSym}}
\newcommand\blfootnote[1]{%
  \begingroup
  \renewcommand\thefootnote{}\footnote{#1}%
  \addtocounter{footnote}{-1}%
  \endgroup
}

\let\oldref\cref
\renewcommand{\cref}[1]{\oldref{#1}}

\newcommand \eq[1]{\begin{equation}\begin{aligned}#1\end{aligned}\end{equation}}
\newcommand \eqtag[2]{\begin{equation}\tag{#1}\begin{aligned}#2\end{aligned}\end{equation}}

\newtheorem{fact}{Fact}
\newtheorem{theorem}{Theorem}

\def\cvprPaperID{11143} 
\def\confName{CVPR}
\def\confYear{2023}

\begin{document}

\title{Semidefinite Relaxations for Robust Multiview Triangulation}
\author{Linus Härenstam-Nielsen$^{1,2}$, Niclas Zeller$^4$, Daniel Cremers$^{1,2,3}$\\
$^1$Technical University of Munich, $^2$Munich Center for Machine Learning, $^3$University of Oxford, \\$^4$Karlsruhe University of Applied Sciences\\
{\tt\small linus.nielsen@tum.de, niclas.zeller@h-ka.de, cremers@tum.de} \\
}
\maketitle

\begin{abstract}
We propose an approach based on convex relaxations for certifiably optimal robust multiview triangulation.  To this end, we extend existing relaxation approaches to non-robust multiview triangulation by incorporating a truncated least squares cost function. We propose two formulations, one based on epipolar constraints and one based on fractional reprojection constraints. The first is lower dimensional and remains tight under moderate noise and outlier levels, while the second is higher dimensional and therefore slower but remains tight even under extreme noise and outlier levels. We demonstrate through extensive experiments that the proposed approaches allow us to compute provably optimal reconstructions even under significant noise and a large percentage of outliers.
\end{abstract}

\section{Introduction}
\label{sec:intro}
\begin{figure}%
    \newcommand{\problemwidth}{\columnwidth}
    \centering
    \begin{subfigure}{\problemwidth}
        \includegraphics[width=\columnwidth, scale=0.01]{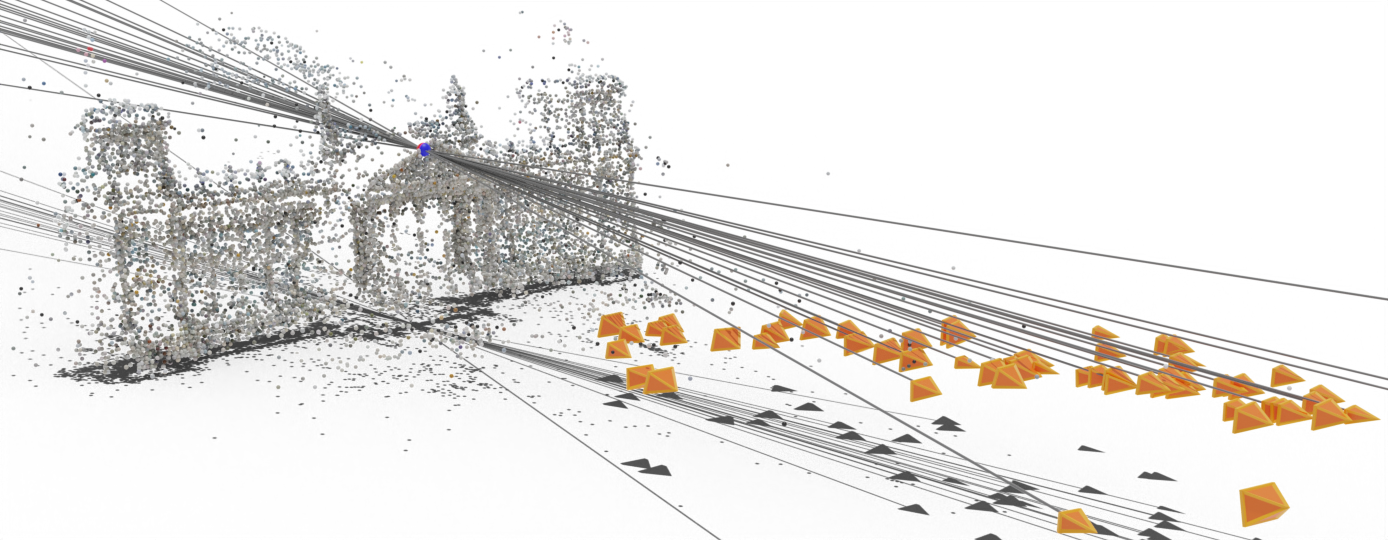}%
        \caption{22 views, no outliers}%
        \label{subfiga}%
    \end{subfigure}\\%
    \begin{subfigure}{\problemwidth}
        \includegraphics[width=\columnwidth]{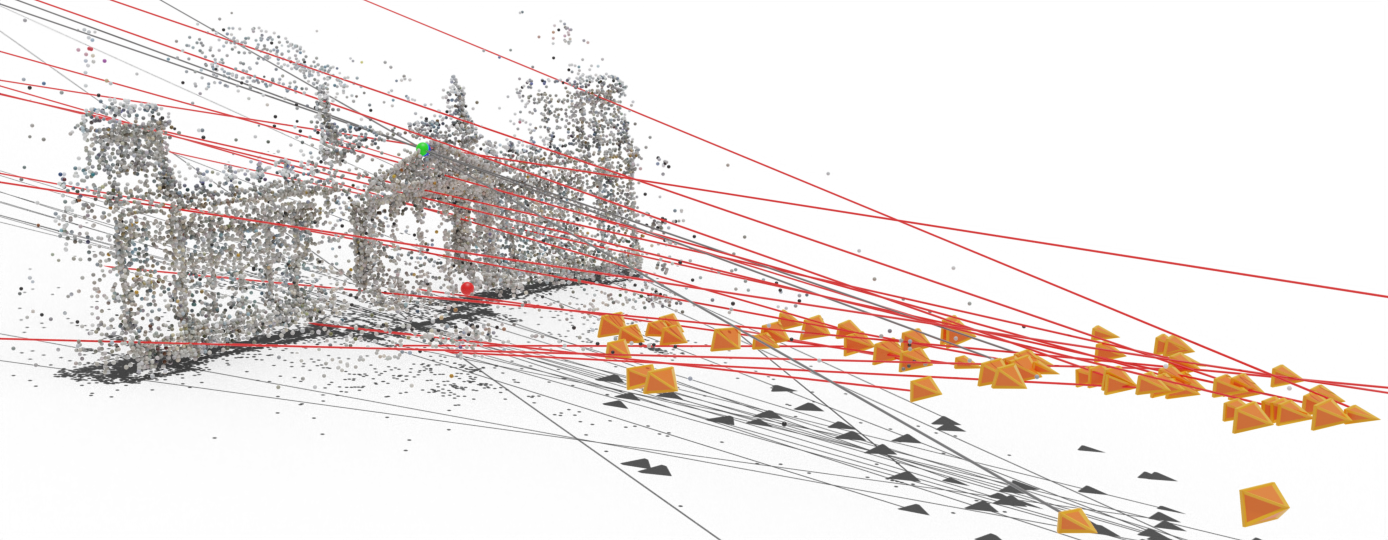}%
        \caption{22 views, 19 outliers}%
        \label{subfigb}%
    \end{subfigure}
    \caption{Example of a triangulated point from the Reichstag dataset. {\color{blue} Blue point}: ground truth from \cite{Jin2020ImageMatching}. {\color{red} Red point}: non-robust global optimum found by the relaxation from \cite{aholt2012qcqp} (see \cref{eqn:triangulation-sdr}). {\color{green} Green point}: robust global optimum found by our proposed relaxation in \cref{eqn:robust-triangulation-relaxation}.}
    \label{fig:reichstag-triangulation}
    \vspace{-0.65cm}
\end{figure}

\blfootnote{\footnotesize{Implementation: \href{https://github.com/linusnie/robust-triangulation-relaxations}{github.com/linusnie/robust-triangulation-relaxations}}}

Multiview triangulation is the problem of estimating the location of a point in 3D given two or more 2D observations in images taken from cameras with known poses and intrinsics.
The 2D observations are typically estimated by some form of feature matching pipeline, so they are always corrupted by noise and outliers. As a result the 3D point cannot be exactly recovered, and instead the solution has to be phrased as a nonconvex optimization problem. 

While solutions are typically computed using faster but sub-optimal local optimization methods, there have also been efforts to compute globally optimal triangulations using semidefinite relaxations \cite{kahl2007globally, aholt2012qcqp, Cifuentes19}. These relaxations can work well even in high-noise scenarios, but their practical use remains limited as they are not robust and even a single outlier can deteriorate the result significantly. In this work, inspired by recent advances in semidefinite relaxations for outlier-robust perception \cite{yang2022certifiably}, we will show that \cite{aholt2012qcqp, Cifuentes19} can be extended to also handle significant amounts of outliers.

Semidefinite relaxations have the advantage of being globally solvable in polynomial time, meaning that they can be used to enable practical \textit{certifiably optimal} algorithms. After solving the relaxed problem we either have that 1) the relaxation is tight and we provably recover the global optimum of the original problem, or 2) the relaxation is provably not tight and we can report failure to find the global optimum. The key metric for the usefulness of a certifiably optimal algorithm is then the percentage of problem cases where the underlying relaxation is tight.

Despite their often slower runtime, certifiably optimal methods offer several advantages:
Firstly, in safety-critical systems it may be required or desirable to complement the computed solution with some guarantee that the solver is not stuck in a local optima. Secondly, in many offline applications runtime is actually not as critical and then one may want to trade off better accuracy for extra runtime. Thirdly, globally optimal solutions of real-world problems can serve as ground truth for assessing the performance of local optimization methods. 

In this work, we demonstrate a certifiably optimal approach to robust triangulation by developing two convex relaxations for the truncated least squares cost function. Enabling the combination of robustness with the capacity to compute certifiably optimal solutions. Our main contributions can be summarized as follows:
\begin{itemize}
    \item We extend the convex triangulation methods from \cite{aholt2012qcqp} and \cite{Cifuentes19} with a truncated least squares cost function and propose two corresponding convex relaxations.
    \item We validate empirically that both relaxations remain tight even under large amounts of noise and high outlier ratios.
    \item We show that the relaxations are tight in the noise-free and outlier-free case by explicitly constructing the dual solution.
\end{itemize}
 To the best of our knowledge, this is the first example of a successful semidefinite relaxation of a robust estimation problem with reprojection errors.

\section{Related work}
Triangulation is a core subroutine for structure from motion and therefore has been studied extensively. For two views, there are many globally optimal solution variants, including computing the roots of a degree 6 polynomial \cite{HARTLEY1997146} for the reprojection error or a $3\times 3$ singular value decomposition for the angular error (up to a second order approximation) \cite{Civera20192viewAngular}. 
Outlier-free multiview triangulation is arguably most commonly solved based on the linear-eigen method from \cite{HARTLEY1997146}.
Robust triangulation is typically tackled using RANSAC \cite{Kang2014PatRec, schoenberger2016sfm, Civera2020RobustMultiViewTriangulation} where a 2-view solver is repeatedly applied to randomly sampled pairs of views until an inlier set can be established.

Semidefinite relaxations have been used to obtain certifiably optimal algorithms for many problems in geometric computer vision.  Examples include semidefinite relaxations for  partitioning, grouping and restoration \cite{keuchel2003binary}, for minimizing reprojection errors \cite{kahl2007globally}, for multiview triangulation \cite{aholt2012qcqp, Cifuentes19}, for essential matrix estimation \cite{zhao2020essential}, for hand-eye calibration \cite{Dekel_2020_CVPR, wise2020handeyeROT, wodtko2021handeyeDQscale}, for robust point cloud registration \cite{Yang2019Wahba, yang2020teaser, yang2022certifiably}, and for 3D shape from 2D landmarks \cite{yang2020_2D3Dshape}. 

Notably \cite{yang2022certifiably}, which is one of the main inpirations for this work, demonstrates that semidefinite relaxations can also be used for outlier-robust estimation. In particular, they provide relaxations for various outliers models in the context of robust rotation averaging, mesh registration, absolute pose registration and category-level object pose+shape estimation. 

Solving semidefinite relaxations is typically slow and memory intensive, stemming from the fact that the number of variables is the square of the number of variables in the original problem. To tackle this issue, there has been recent interest in developing solvers that can scale to larger problems. Including \cite{dathathri2020eigenvalueSDP} which uses a reformulation in terms of eigenvalue optimization based on \cite{helmberg2000spectral} which can take advantage of GPUs, and \cite{yang2022certifiably} which uses efficient non-global solvers for speeding up the convergence of the global solver. Notably, in both cases the main memory saving comes from applying a dual-only solver.

In a limited number of cases, semidefinite relaxations can be shown to always find a globally optimal solution to the original problem. This includes the dual quaternion formulation of hand-eye calibration \cite{Dekel_2020_CVPR} and 2-view triangulation using epipolar constraints \cite{aholt2012qcqp}, in both cases with some assumption of non-degenerate measurements. Another example is the rotation alignment problem which has a closed form solution in terms of an eigenvalue decomposition (quaternion formulation) or singular value decomposition (rotation matrix formulation).

However, outlier-robust estimation is inapproximable in general \cite{yang2022hardness}, meaning there will always be some subset of possible measurements for which any semidefinite relaxation of practical size is non-tight. 
In terms of theoretical guarantees, \cite{cifuentes2022local} introduces the concept of \textit{local stability} which, under certain conditions on the problem structure for noise-free measurements, can guarantee that a relaxation remains tight for bounded measurement noise.
In some cases it is also possible to find conditions on measurements which guarantee that the relaxation is tight or non-tight, as demonstrated in \cite{peng2022semidefinite} for robust rotation alignment of point clouds, but typically algorithm developers will have to rely on experiments in order to determine to what extent a given relaxation remains tight in a particular problem scenario. 

\section{Notation and preliminaries}
\label{sec:notation}
For $t, s \in \mathbb{R}^3$ we write $[t]_{\times}$ for the $3\times 3$ skew-symmetric matrix such that $t\times s = [t]_\times s$. $(a\cat b)$ denotes the vertical concatenation of vectors $a$ and $b$ and for a collection of vectors $a_1, \ldots, a_n$ the subscript-free version denotes the corresponding stacked vector $a = (a_1 \cat\ldots\cat a_n)$. We use a bar to denote the \textit{homogeneous} version of a vector, that is $\bar a \coloneqq (a; 1)$. When dimensionality is understood we define $e_i$ to be the $i$th unit vector and $E_i = e_ie_i^T$. For a vector of monomials $m=(m_1\cat \ldots\cat m_d)$ we define $e^m_{m_i}$ as the unit vector whose only non-zero entry corresponds to the index of $m_i$ in $m$, meaning $e^m_{m_i} = e_i \in\mathbb{R}^d$. For a vector $x\in\mathbb{R}^k$ we define:
\eq{
    M_x \coloneqq \begin{pmatrix}
            I & -x
        \\  -x^T & \|x\|^2
    \end{pmatrix}\in\psdset{k+1}
}
such that for $y\in\mathbb{R}^k$ we have $\bar y^TM_x\bar y = \|x - y\|^2$. The operator $\otimes$ denotes the Kronecker product, and $\oplus$ denotes the tensor sum. For example, for $2\times 2$ matrices $A$ and  $B$:
\eq{
    A\oplus B = \begin{pmatrix}
        A & 0
    \\  0 & B
    \end{pmatrix}, A\otimes B = \begin{pmatrix}
        a_{11}B & a_{12}B
    \\  a_{21}B & a_{22}B
    \end{pmatrix}.
}

\subsection{Semidefinite relaxations}
\label{sec:semidefinite-relaxations}
As a general strategy, we aim to solve the triangulation problem by relaxing a \textit{Quadratically Constrained Quadratic Program} (QCQP) which has the following form:
\eq{
\label{eqn:qcqp}
    \min_{z\in\mathbb{R}^d} \quad & z^TMz
    \\ \textrm{s.t.} \quad & z^TEz = 1
    \\ & z^TA_iz = 0, \quad i=1,\ldots,k
    \\ & z^TB_jz \leq 0, \quad j=1,\ldots,l.
}
This is a very general formulation with applications in computer vision but it is NP-hard to solve in most cases, so an imperfect method is typically necessary.  One such strategy is to lift the problem from $\mathbb{R}^d$ to the set of $d\times d$ positive semidefinite matrices, $\psdset{d}$, by introducing a new variable $Z = zz^T$ and using the fact that $z^TMz = \tr(Mzz^T) = \tr(MZ)$ to arrive at:
\eq{
\label{eqn:primal}
    \min_{\psd\in \psdset{d}} \quad & \tr(M\psd)
    \\ \textrm{s.t.} \quad & \tr(E\psd)=1
    \\ & \tr(A_i\psd) = 0, \quad i=1,\ldots,k
    \\ & \tr(B_j\psd) \leq 0 \quad j=1,\ldots, l
}
\cref{eqn:primal} is a relaxation of \cref{eqn:qcqp} since if $z$ satisfies the constraints of \cref{eqn:qcqp} we always have that $\psd = zz^T$ satisfies the constraints of \cref{eqn:primal} with the same objective value. However, the converse is not always true. In particular, if $\hat Z$ is optimal for \cref{eqn:primal} we can obtain a corresponding solution $\hat z$ for \cref{eqn:qcqp} with the same objective value if and only if $\hat Z$ is rank one. In this case we have $\hat Z = \hat z\hat z^T$ and we then say that the relaxation is \textit{tight}. 

The main advantage of working with the relaxation \cref{eqn:primal} as opposed to the original problem \cref{eqn:qcqp} is that the relaxation is a convex optimization problem, in particular it is a semidefinite program, for which a variety of polynomial-time solvers are available, including \cite{mosek, Boyd2016SCS}. 
If the relaxation is not tight we can at best expect an optimal $\hat Z$ to generate an approximation of the optimal $\hat z$. Therefore, a key metric to consider when applying a relaxation is the percentage of encountered problem cases in which it remains tight.

\section{Relaxations for multiview triangulation}
\begin{figure*}%
    \newcommand{\problemwidth}{.6\columnwidth} 
    \newcommand{\trim}{10cm}
    \centering
    \begin{subfigure}{\problemwidth}
        \includegraphics[trim={0 0 0 \trim}, clip, width=\columnwidth]{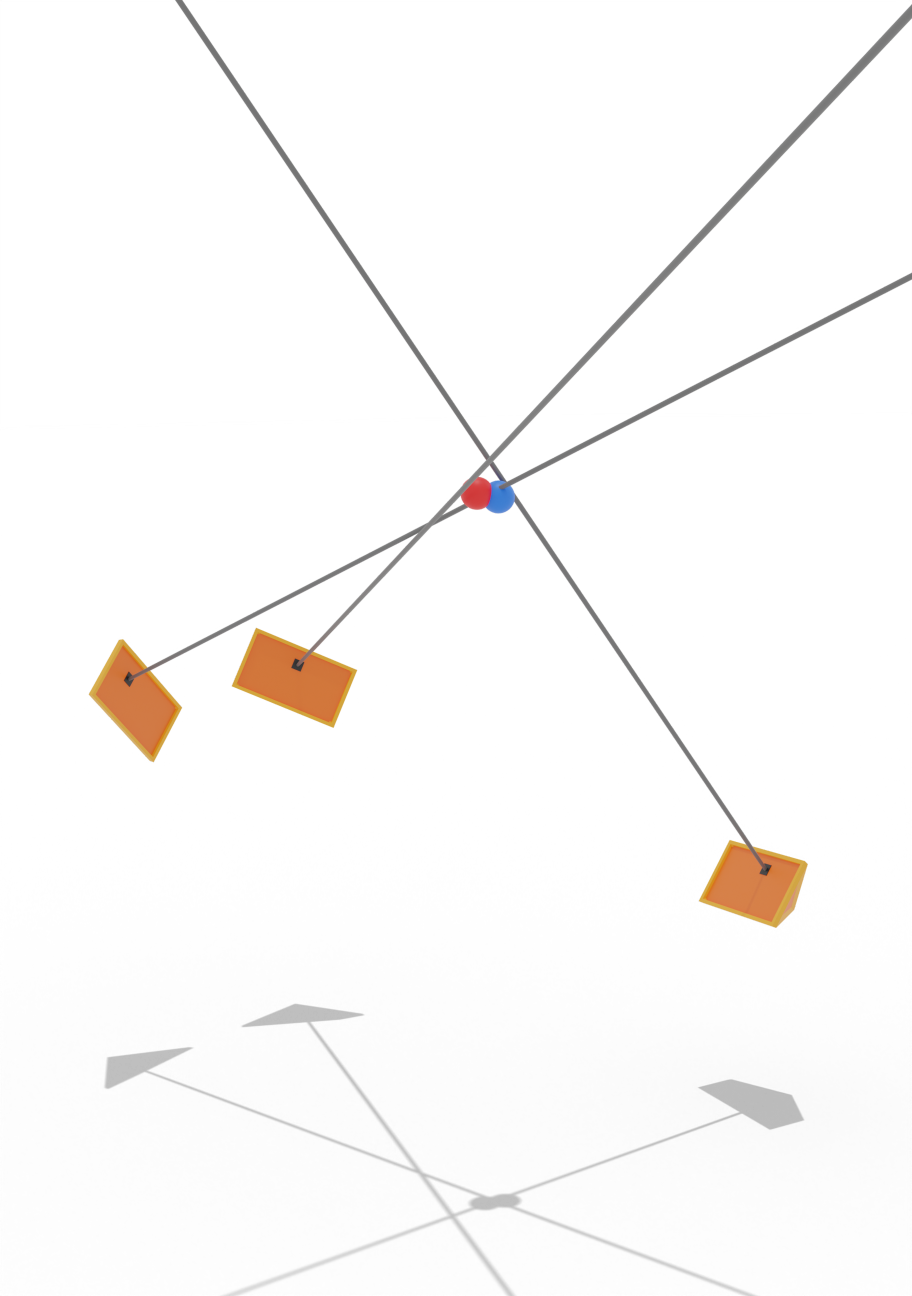}%
        \caption{3 views}%
        \label{subfiga}%
    \end{subfigure}\quad%
    \begin{subfigure}{\problemwidth}
        \includegraphics[trim={0 0 0 \trim}, clip, width=\columnwidth]{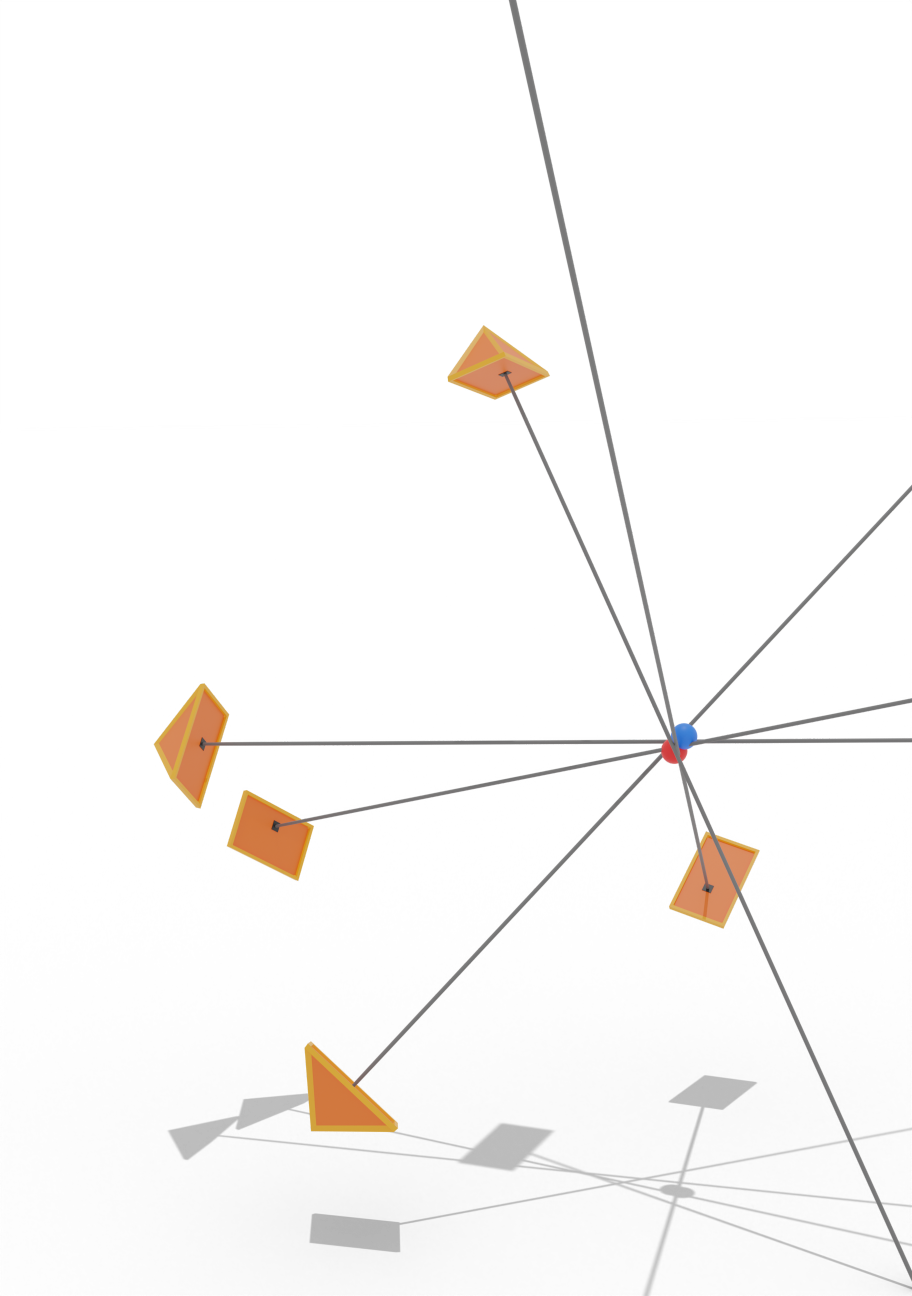}%
        \caption{5 views}%
        \label{subfigb}%
    \end{subfigure}\quad%
    \begin{subfigure}{\problemwidth}
        \includegraphics[trim={0 0 0 \trim}, clip, width=\columnwidth]{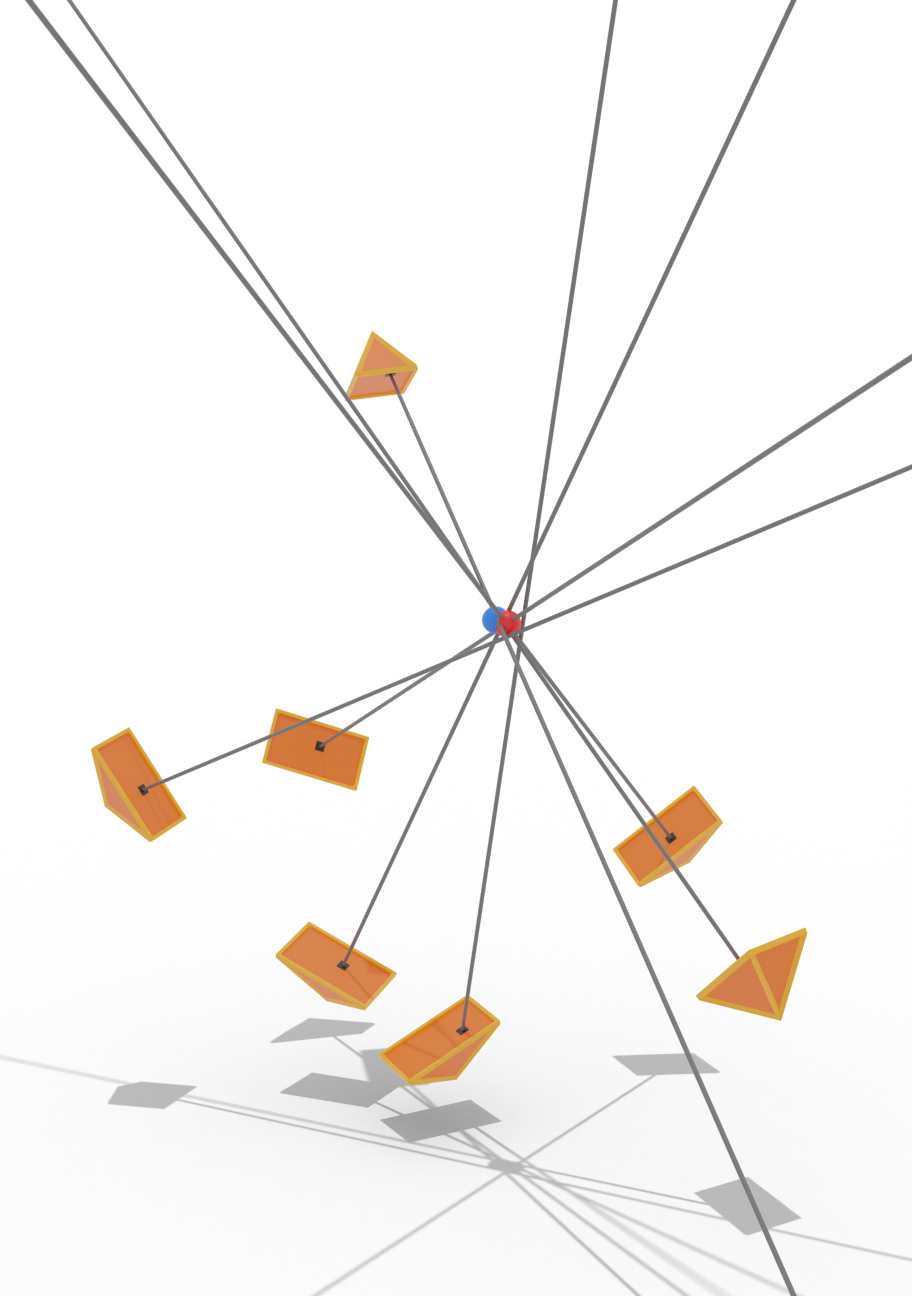}%
        \caption{7 views}%
        \label{subfigb}%
    \end{subfigure}\hfill%

    \begin{subfigure}{\problemwidth}
        \includegraphics[trim={0 0 0 \trim}, clip, width=\columnwidth]{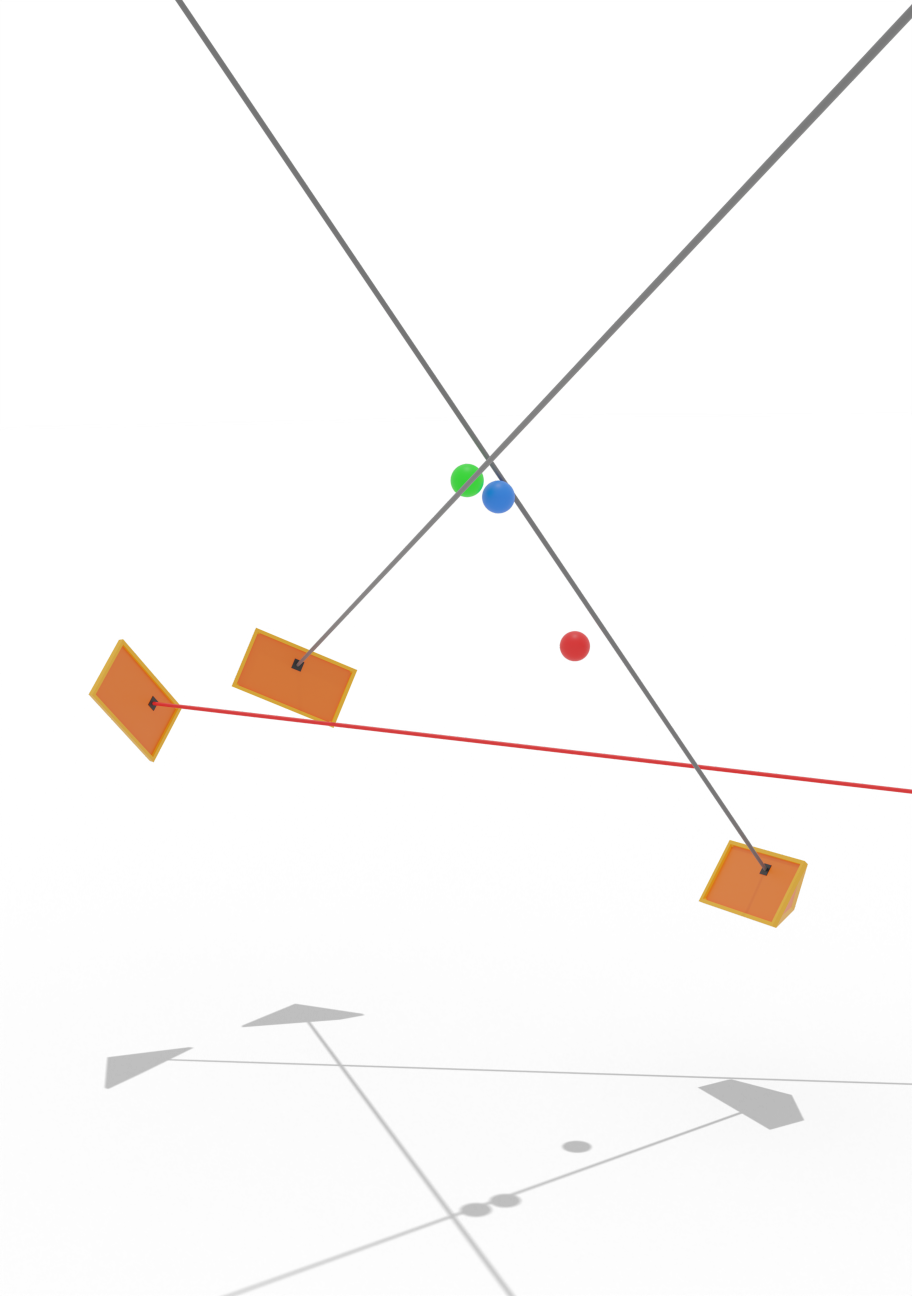}%
        \caption{3 views, 1 outlier}%
        \label{subfiga}%
    \end{subfigure}\quad%
    \begin{subfigure}{\problemwidth}
        \includegraphics[trim={0 0 0 \trim}, clip, width=\columnwidth]{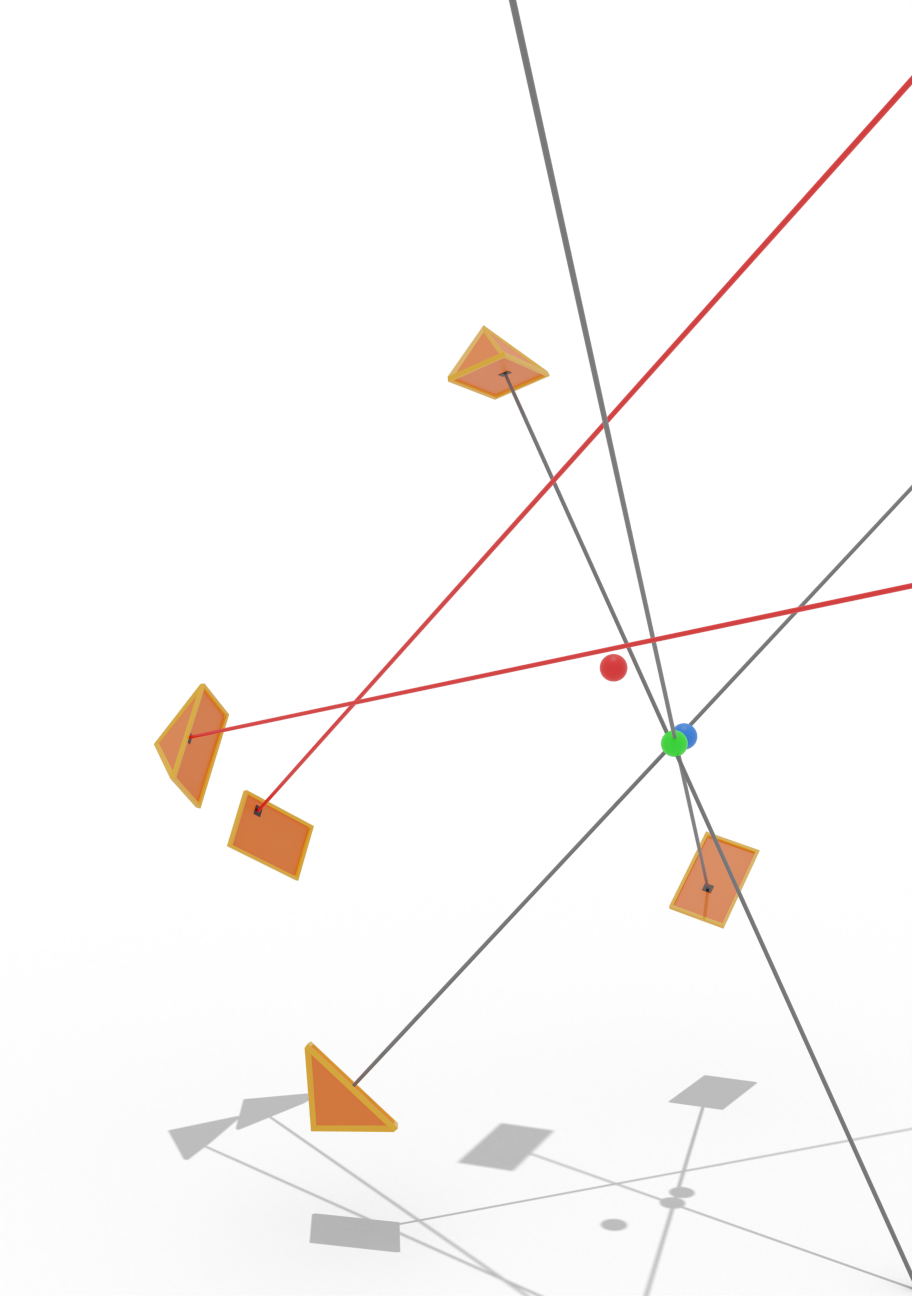}%
        \caption{5 views, 2 outliers}%
        \label{subfigb}%
    \end{subfigure}\quad%
    \begin{subfigure}{\problemwidth}
        \includegraphics[trim={0 0 0 \trim}, clip, width=\columnwidth]{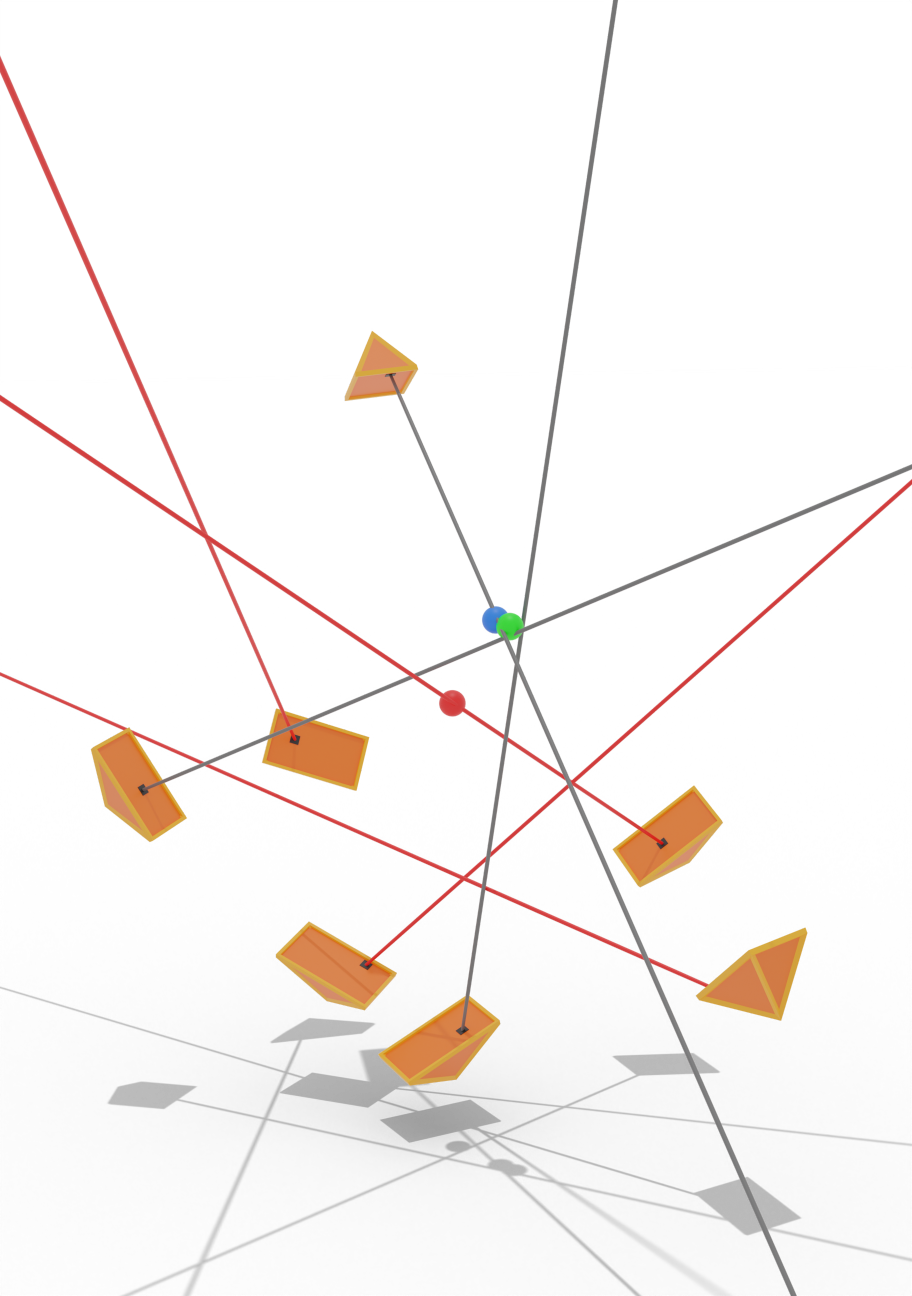}%
        \caption{7 views, 4 outliers}%
        \label{subfigb}%
    \end{subfigure}\hfill%
    \caption{Examples of simulated triangulation problems from \cref{sec:synthetic} with $\sigma = 50$px for various number of views and outliers.  {\color{blue} Blue point}: ground truth, {\color{red} Red point}: non-robust global optimum found by the relaxation from \cite{Cifuentes19} (see \cref{eqn:triangulation-fractional-relaxation}). {\color{green} Green point}: robust global optimum found by our proposed relaxation in \cref{eqn:robust-triangulation-fractional-relaxation}. With no outliers the robust and non-robust methods give the same result.}
     \vspace{-0.55cm}
    \label{fig:triangulation-problems}
\end{figure*}

\noindent Given $n$ views of a point $X$ from cameras located at $P_i = (R_i, t_i)\in\text{SE}(3)$ in camera-to-world convention with intrinsic matrices $K_i\in\mathbb{R}^{3\times 3}$, and with, possibly noisy, observations denoted as $\obs_i\in\mathbb{R}^2$, the $n$-view triangulation problem with reprojection error is defined as:

\eq{
\label{eqn:triangulation}
    \min_{\point\in\mathbb{R}^3} \sum_{i=1}^{n}\|\obs_i - \pi(K_i, P_i, X)\|^2
}
where $\pi(K_i, P_i, \point)$ is the reprojection of the point $\point\in\mathbb{R}^3$ to camera $i$. This is a nonconvex problem but it is not yet in QCQP form since $\pi(K_i, P_i, X)$ is not quadratic in $X$. In this section we will recap two ways of converting \cref{eqn:triangulation} to a QCQP, from which we can generate the corresponding semidefinite relaxations.

\subsection{Triangulation with epipolar constraints}
The first approach was introduced in \cite{aholt2012qcqp}. They showed that \cref{eqn:triangulation} can be formulated as a polynomial optmization problem of degree 2 by reparametrizing $\point$ in terms of it's $n$ reprojections $\repr_i$, which are constrained to satisfy the epipolar constraints:
\eq{
\label{eqn:triangulation-poly}
    \min_{\repr_i\in\mathbb{R}^2} \quad & \sum_{i=1}^{n}\|\repr_i - \tilde \repr_i\|^2
    \\ \textrm{s.t.} \quad & \homrepr_i^TF_{ij}\homrepr_{j} = 0
    \\ & i, j = 1,\ldots,n \quad i\neq j
}
where $F_{ij} = K_i^{-T}[t_{ij}]_\times R_{ij}K_j^{-1}$ is the fundamental matrix corresponding to the relative transformation between poses $i$ and $j$. Since the estimated reprojections $x_i$ all satisfy the epipolar constraints, the solution of \cref{eqn:triangulation} can in most cases be recovered exactly from \cref{eqn:triangulation-poly} using the linear-eigen method from \cite{HARTLEY1997146}. An important failure case of this parametrization is that when the camera centers are co-planar it is possible that the solution $x_i$ of \cref{eqn:triangulation-poly} does not correspond to a valid 3D point, see \cref{sec:coplanar} for more details and an example.

Using the parametrization $z=(x\cat 1)=\homrepr$ the semidefinite relaxation of \cref{eqn:triangulation-poly} is:
\eqtag{T}{
\label{eqn:triangulation-sdr}
    \min_{\psd\in\psdset{2n+1}} \quad & \tr(M_{\tilde x}\psd)
    \\ \textrm{s.t.} \quad & \tr(E_{n+1}\psd)=1
    \\ & \tr(\bar F_{ij}\psd) = 0, \quad i=1,\ldots,k
}
where $\bar F_{ij}$ is a symmetric $(2n+1)\times(2n+1)$ matrix defined such that $\homrepr^T \bar F_{ij} \homrepr = \homrepr_i^TF_{ij}\homrepr_j$.
It was shown in both \cite{aholt2012qcqp} and \cite{cifuentes2022local} that the relaxation \cref{eqn:triangulation-sdr} is locally stable with respect to noise as long as the views are not co-planar. 

\subsection{Triangulation with fraction constraints}
As initially proposed in \cite{Cifuentes19}, an alternative to \cref{eqn:triangulation-poly} is to explicitly parametrize the 3D point $X$ in homogeneous coordinates. This leads to the fractional reprojection constraints $x_i^k = a_{ik}^T\hompoint/b_i^T\hompoint$, where $a_{i1}$, $a_{i2}$ and $b_i$ are given by the rows of the $i$-th camera projection matrix $K_i\begin{pmatrix} R_i^T & -R_i^Tt_i \end{pmatrix}$. By multiplying out the right hand side denominators we get the following QCQP:
\eq{
\label{eqn:triangulation-fractional}
    \min_{\hompoint\in\mathbb{R}^4, x_i\in\mathbb{R}^2} \quad & \sum_{i=1}^{n}\|\repr_i - \obs_i\|^2
    \\ \textrm{s.t.} \quad & \hompoint^T\hompoint = 1
    \\ & \repr_i^kb_{i}^T\hompoint - a_{ik}^T\hompoint = 0
    \\ & i=1,\ldots,n\quad k=1, 2.
}
A naive approach to relaxing \cref{eqn:triangulation-fractional} would be to use the parametrization $z = (\repr\cat\hompoint)$, but as shown in \cite{Cifuentes19} this leads to a relaxation whose optimal value is always zero. To circumvent this issue, \cite{Cifuentes19} instead proposes parametrizing the problem in terms of all possible products between the elements of $\repr$ and $\point$, \ie $\repr\otimes\point$. They also show through experiments that, while the resulting relaxation has more parameters and constraints than \cref{eqn:triangulation-sdr}, it is also tight in a significantly wider range of cases, leading to a tradeoff between stability and computation time.

We will use a similar relaxation, though we will skip the initial change of variables to get a slightly different but equivalent formulation which can be extended to the robust case more conveniently. We start by setting $z = (\repr\otimes \hompoint\cat\hompoint) = \homrepr\otimes\hompoint$ and then we multiply each reprojection constraint in \cref{eqn:triangulation-fractional} with $z_j$ to get $8n+4$ quadratic constraints:
\eq{
    \label{eqn:reprojection-constraints-zj}
    (\repr_i^kb_{i}^T\hompoint - a_{ik}^T\hompoint)z_j 
    &= z^T(e^{\homrepr}_{x_i^k}\otimes b_{i} - e^{\homrepr}_1\otimes a_{ik})e_j^Tz 
    \\ &= 0
}
where we have made use of the unit vector notation from \cref{sec:notation}, meaning in particular $e^{\homrepr}_{x_i^k} = e_{2i + k}$ and $e^{\homrepr}_{1} = e_{2n + 1}$. We also need to introduce constraints to preserve the fact that $z$ comes from a $(2n+1)\times 4$ Kronecker product. When $Z=zz^T$ is rank one, it turns out that this condition is equivalent to $Z$ being composed of $2n+1$ symmetric $4\times 4$ blocks, see \cite{Cifuentes19} for more details. We will denote this constraint as $Z\in\SM(2n+1, 4)$. The relaxation of can then be written as\footnote{The cost functions in \cref{eqn:triangulation-fractional} and \cref{eqn:triangulation-fractional-relaxation} are equivalent, since $(\hompoint\otimes\homrepr)^T(M_{\tilde x}\otimes I_4)(\hompoint\otimes\homrepr) = (\homrepr^TM_{\tilde x}\homrepr)\hompoint^T\hompoint = \homrepr^TM_{\tilde x}\homrepr$.}:
\eqtag{TF}{
\label{eqn:triangulation-fractional-relaxation}
    \span\min_{\substack{
        \psd\in\mathbb{S}_+^{8n + 4}
    }} 
    \quad \tr(\psd(M_{\obs}\otimes I_4))
    \\ \textrm{s.t.} \quad & \tr(\psd (0_{8n\times 8n}\oplus I_4)) = 1
    \\ & \psd\in\SM(2n+1, 4)
    \\ & \tr(\psd(e^{\homrepr}_{x_i^k}\otimes b_{i} - e^{\homrepr}_1\otimes a_{ik}) e_j^T) = 0
    \\ & i=1,\ldots n,\quad k=1, 2
    \\ & j=1,\ldots,8n+4.
}
\section{The robust case}
Now that we have introduced the two main relaxations of \cref{eqn:triangulation} we move to the the main contribution of this paper, which is to introduce the corresponding truncated least squares (TLS) extensions. Similarly to \cite{yang2022certifiably} we will use the fact that the TLS cost function can be written as a minimization problem by introducing a binary decision variable for each residual 
\eq{
\label{eqn:robust-residual}
    \rho_i(r_i^2)
    = \min(r_i^2, c_i)
    = \min_{\theta_i\in\{0, 1\}}{\theta_i r_i^2 + (1 - \theta_i)c_i}
}
where $c_i > 0$ is the square of the inlier threshold. In \cref{sec:theory} we show that both relaxations are tight in the noise-free and outlier-free case, and we also show part of the criteria required for local stability.

\subsection{Robust triangulation with epipolar constraints}
Using \cref{eqn:robust-residual}, the TLS extension of \cref{eqn:triangulation-poly} is:
\eq{
\label{eqn:robust-triangulation-order2}
    \min_{x_i\in\mathbb{R}^2,\theta_i\in\mathbb{R}} \quad & \sum_{i=1}^{n}\Bigg(\theta_i\|\repr_i - \obs_i\|^2 
        + (1-\theta_i) c_i\Bigg)
    \\ \textrm{s.t.} \quad & \homrepr_i^TF_{ij}\homrepr_j = 0,
    \\ & \theta_i^2 - \theta_i = 0, \quad\textstyle\sum_{i=1}^n\theta_i^2 \geq 2,
    \\ & i, j = 1,\ldots,n \quad i\neq j.
}
where the constraints $\theta_i^2 - \theta_i = 0$ ensures $\theta_i$ equals 0 or 1, and the constraint $\sum_{i=1}^n\theta_i^2\geq 2$ ensures there are at least two inliers in the final solution. This cost function includes terms like $\theta_i\|x_i\|^2$, which means it is a 3rd degree polynomial, so we can't apply the relaxation as described in \cref{sec:semidefinite-relaxations} directly. But we can obtain a 2nd order formulation by noting that $\theta_i^2 = \theta_i$ implies $\theta_i\|x_i - \obs_i\|^2 = \|\theta_ix_i - \theta_i \obs_i\|^2$ and making the substitution $y_i=\theta_ix_i$:
\eq{
\label{eqn:robust-triangulation-order2-y}
    \min_{y_i\in\mathbb{R}^2,\theta_i\in\mathbb{R}} \quad & \sum_{i=1}^{n}\Bigg(\|y_i - \theta_i\obs_i\|^2 
        +(1-\theta_i)c_i\Bigg)
    \\ \textrm{s.t.} \quad & (y_i\cat\theta_i)^TF_{ij}(y_j\cat\theta_j) = 0
    \\ & \theta_i^2 - \theta_i = 0, \quad\textstyle\sum_{i=1}^n\theta_i^2 \geq 2,
    \\ & \theta_iy_i = y_i
    \\ & i,j = 1,\ldots,n, \quad i\neq j.
}
The last set of constraints $\theta_iy_i=y_i$ are redundant but we find that they are necessary for the relaxation to remain tight in the presence of noise, these are referred to as moment constraints in \cite{yang2022certifiably}. We can recover a solution of \cref{eqn:robust-triangulation-order2} from a solution of \cref{eqn:robust-triangulation-order2-y} by triangulating the estimated inliers (\ie $y_i$ for which $\theta_i = 1$) and setting each $x_i$ (including outliers) to be the reprojection of the resulting point onto view $i$.

Using the parametrization $z = (y\cat\theta\cat 1)$ the semidefinite relaxation of \cref{eqn:robust-triangulation-order2-y} is:
\eqtag{RT}{
\label{eqn:robust-triangulation-relaxation}
    \min_{\psd\in S_{+}^{3n+1}} \quad & \ \text{tr}(M_{\obs}^c\psd)
    \\ \textrm{s.t.} \quad & \\ & \tr(E_{3n+1}Z) = 1
    \\ & \text{tr}(\bar F_{ij}\psd) = 0
    \\ & Z_{\theta_i, \theta_i} - Z_{1, \theta_i} = 0
    \\ & Z_{\theta_i, y_i} - Z_{1, y_i} = 0
    \\ & \textstyle\sum_{i=1}^nZ_{\theta_i,\theta_i}\geq 2Z_{1,1}
    \\ & i, j = 1,\ldots,n \quad i\neq j
}
where $M_{\obs}^c$ is the robust extension of $M_{\obs}$, defined as:
\eq{
    &M_{\obs}^c = \begin{pmatrix}
        I & -B(\obs) & 0
    \\  -B(\obs)^T & \text{diag}(\|\obs_i\|^2) & -c
    \\ 0 & -c^T & \sum_{i=0}^nc_i
    \end{pmatrix},
    \\&B(\tilde x) = \begin{pmatrix}
        \obs_1 & 0 & \ldots & 0
    \\  0 & \obs_2 & \ldots & 0
    \\ \vdots & \vdots & \ddots & 0
    \\ 0 & 0 & 0 & \obs_n
    \end{pmatrix}.
}
and $Z_{m_i, m_j}$ is the entry of $Z$ corresponding to the index of the monomials $m_i$ and $m_j$ in $z$. In \cref{sec:local-stability-epipolar} we analyze the dual of \cref{eqn:robust-triangulation-relaxation} to show that the relaxation is tight in the noise-free and outlier-free case. 

\subsection{Robust triangulation with fraction constraints}
\label{sec:robust-fractional}
In this section we will introduce a higher order relaxation which can handle higher noise and outlier levels.
Since the fractional constraints in \cref{eqn:triangulation-fractional-relaxation} are more stable with respect to noise than the epipolar constraints in \cref{eqn:triangulation-sdr}, we might also expect that extending \cref{eqn:triangulation-fractional-relaxation} to handle outliers will result in a relaxation which is more stable than \cref{eqn:robust-triangulation-relaxation}. In this section we will show how the robust extension can be formulated, and as we will see in \cref{sec:experiments} it is indeed significantly more stable with respect to both noise and outliers.

In order to extend \cref{eqn:triangulation-fractional-relaxation} to handle outliers we will proceed in a similar manner as in the case with epipolar constraints. Starting by writing the cost function in terms of the 2nd order variables $y_i = \theta_i x_i$:
\eq{
\label{eqn:robust-triangulation-fractional}
    \min_{\substack{\hompoint\in\mathbb{R}^4, y_i\in\mathbb{R}^2 \\ \theta_i\in\mathbb{R}}} \quad & \sum_{i=1}^{n}\Bigg(\|y_i - \theta_i\obs_i\|^2 + (1 - \theta_i)c_i\Bigg)
    \\ \textrm{s.t.} \quad & \hompoint^T\hompoint = 1
    \\ & y_i^kb_{i}^T\hompoint - \theta_ia_{ik}^T\hompoint = 0
    \\ & \theta_i^2 - \theta_i = 0, \quad\textstyle\sum_{i=1}^n\theta_i^2 \geq 2
    \\ & \theta_i y_i = y_i
    \\ & i=1,\ldots,n\quad k=1, 2.
}
For convenience we will denote the vertical concatenation of $y$ and $\theta$ as $(y\cat\theta) = y_\theta$. For the relaxation we will then use the parametrization $z = (y_\theta\otimes\hompoint\cat \hompoint) = \bar y_{\theta}\otimes\hompoint$ and generate redundant moment constraints from $\theta_i^2-\theta_i = 0$ and $\theta_iy_i = y_i$ by multiplying each equation by $\hompoint_s\hompoint_t$ for $s, t = 1,\ldots,4$. We also generate redundant inequalities from $\sum_{i=1}^n\theta_i^2 \geq 2$ by multiplying by $\hompoint_s^2$ for each $s=1,\ldots, 4$. Resulting in the following relaxation:

\eqtag{RTF}{
\label{eqn:robust-triangulation-fractional-relaxation}
     \span\min_{\substack{
        \psd\in\mathbb{S}_+^{12n + 4}
    }} 
    \quad \tr(\psd((M_{\obs}^c\otimes I_4)\oplus 0_{4\times 4}))
    \\ \textrm{s.t.} \quad & \tr(\psd (0_{12n\times 12n}\oplus I_4)) = 1
    \\ & \psd\in\SM(3n+1, 4)
    \\ & \tr(\psd(e_{y_i^k}^{\bar y_\theta}\otimes b_{i} - e_{\theta_i}^{\bar y_\theta}\otimes a_{ik}) e_j^T) = 0
    \\ & Z_{\hompoint_s\theta_i, \hompoint_t\theta_i} - Z_{\hompoint_s, \hompoint_t\theta_i} = 0
    \\ & Z_{\hompoint_s\theta_i, \hompoint_ty_i} - Z_{\hompoint_s, \hompoint_ty_i} = 0
    \\ & \textstyle\sum_{i=1}^n\psd_{\theta_i\hompoint_s, \theta_i\hompoint_s} \geq 2\psd_{\hompoint_s, \hompoint_s}
    \\ & i=1,\ldots n, \quad k=1, 2
    \quad s, t = 1, \ldots, 4
    \\ & j=1,\ldots,12n+4.
}
Similarly to the epipolar case we can show that this relaxation is tight in the noise-free and outlier-free case by explicitly constructing the globally optimal Lagrange multipliers, see \cref{sec:local-stability-fractional} for details.

With this we have 4 relaxations for the triangulation problem corresponding to the non-robust and robust case with the epipolar and the fractional parametrization. We summarize the relaxations and their number of variables and constraints in \cref{tab:relaxations}.
\begin{table}
    \resizebox{\columnwidth}{!}{\begin{tabular}{llcll}
        \midrule
        Problem & Relaxation & Robust & Constraints & Variables 
        \\ \hline
        \cref{eqn:triangulation-poly} & \cref{eqn:triangulation-sdr} & \xmark & $\frac{1}{2}n^2-\frac{1}{2}n + 1$ & $2n+1$
        \\ 
        \cref{eqn:robust-triangulation-order2-y} & \cref{eqn:robust-triangulation-relaxation} & \cmark & $\frac{1}{2}n^2 + 2.5n + 1$ & $3n+1$
        \\
        \cref{eqn:triangulation-fractional} & \cref{eqn:triangulation-fractional-relaxation} & \xmark & $28n^2 + 14n$ + 1 & $8n + 4$
        \\
        \cref{eqn:robust-triangulation-fractional} & \cref{eqn:robust-triangulation-fractional-relaxation} & \cmark & $51n^2 + 65n + 1$ & $12n + 4$
        \\ \hline
        \bottomrule
    \end{tabular}}
        \vspace{-0.25cm}
        \caption{Summary of relaxations for the triangulation problem and its robust extension.}
        \label{tab:relaxations}
         \vspace{-0.65cm}
\end{table}
\begin{figure*}
    \begin{center}
        \vspace{-0.5cm}
        \includegraphics[width=\linewidth]{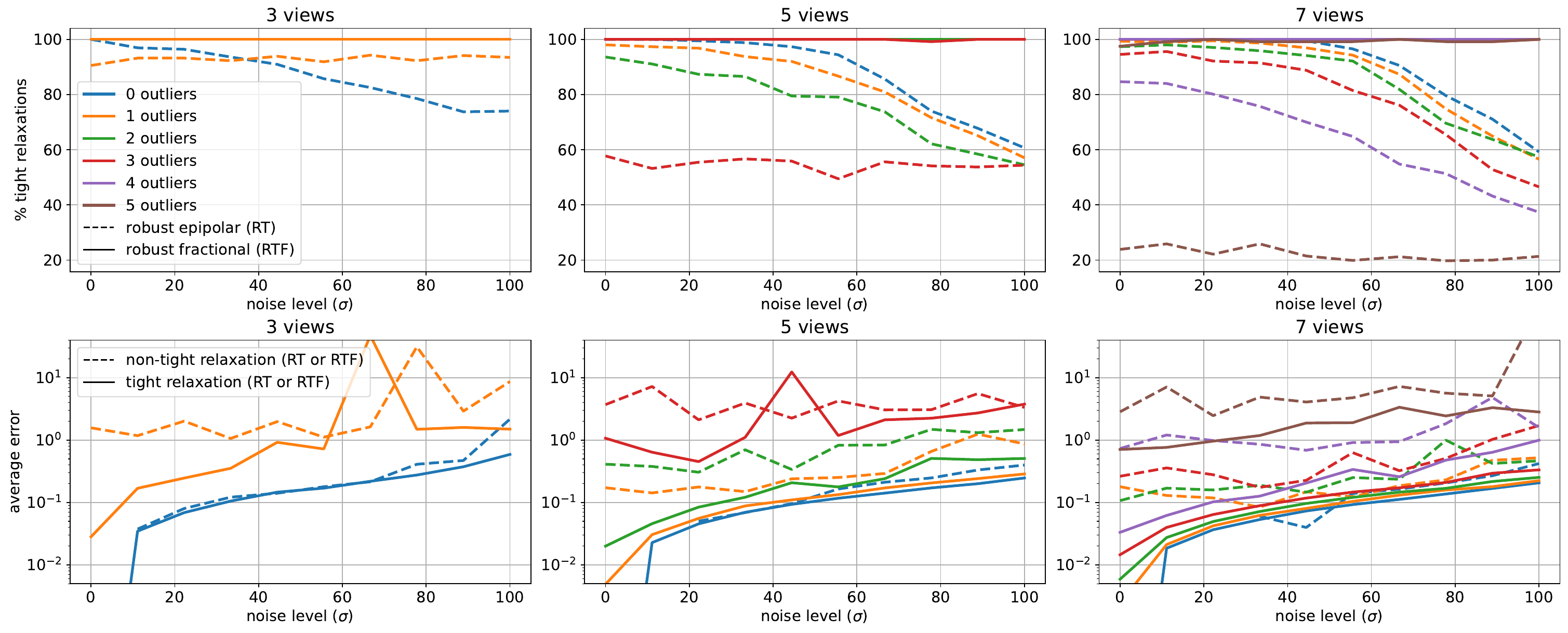} 
    \end{center}
    \vspace{-0.75cm}
    \caption{Average number of tight relaxation (top) and estimation error (bottom) for 3, 5 and 7 views for the robust epipolar relaxation \cref{eqn:robust-triangulation-relaxation} and the robust fractional relaxation \cref{eqn:robust-triangulation-fractional-relaxation} for experiments described in \cref{sec:synthetic}. Note that the error plot is divided by tight/non-tight relxation since both methods give the same result whenever the relaxation is tight.}
    \label{fig:experiments-error-tightness}
\end{figure*}

\subsection{Rounding in the non-tight case}
For non-tight cases the optimal $\hat Z$ will have rank of at least 2, which means we can't recover the optimal solution $\hat z$ for the original problem \cref{eqn:qcqp}. However we can still construct an approximate solution through a rounding procedure. We start by setting $\hat z$ to be the eigenvector corresponding to the minimal eigenvalue, normalized such that $\hat z^TE\hat z = 1$. We then apply a different procedure for each problem depending on the constraints. For \cref{eqn:triangulation-sdr} we triangulate the resulting $\hat x_i$ (which in this case will generally not satisfy the epipolar constraints) using the linear-eigen method from \cite{HARTLEY1997146}. For \cref{eqn:robust-triangulation-relaxation} we do the same except that we first determine the inlier parameters $\hat \theta_i$ by rounding the corresponding entries of $\hat z$ to 0 or 1. For \cref{eqn:triangulation-fractional-relaxation} and \cref{eqn:robust-triangulation-fractional-relaxation} we compute the best-fitting tensor product decomposition of $\hat z$ using a singular value decomposition as described in \cite{VanLoan1993KroneckerDecomp} and then use the same method as in the epipolar case for determining the inlier parameters. 

\begin{figure}
    \begin{center}
        \vspace{-0.5cm}
        \includegraphics[width=\linewidth]{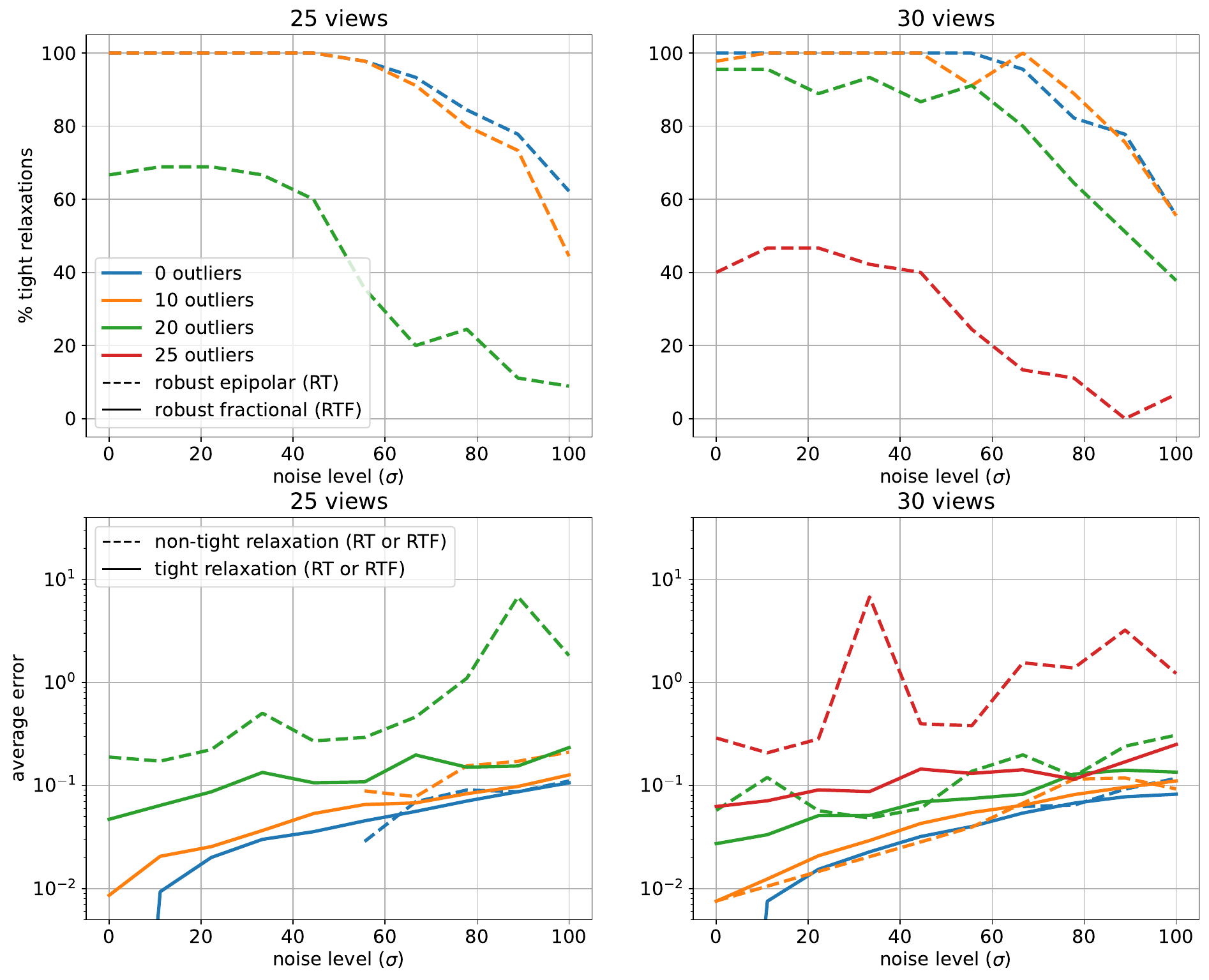} 
    \end{center}
    \vspace{-0.75cm}
    \caption{Average number of tight relaxation (top) and estimation error(bottom) for 25 and 30 views using \cref{eqn:robust-triangulation-relaxation}.}
    \label{fig:experiments-error-tightness-25}
\end{figure}
\begin{figure}
    \begin{center}
        \vspace{-0.5cm}
        \includegraphics[width=\linewidth]{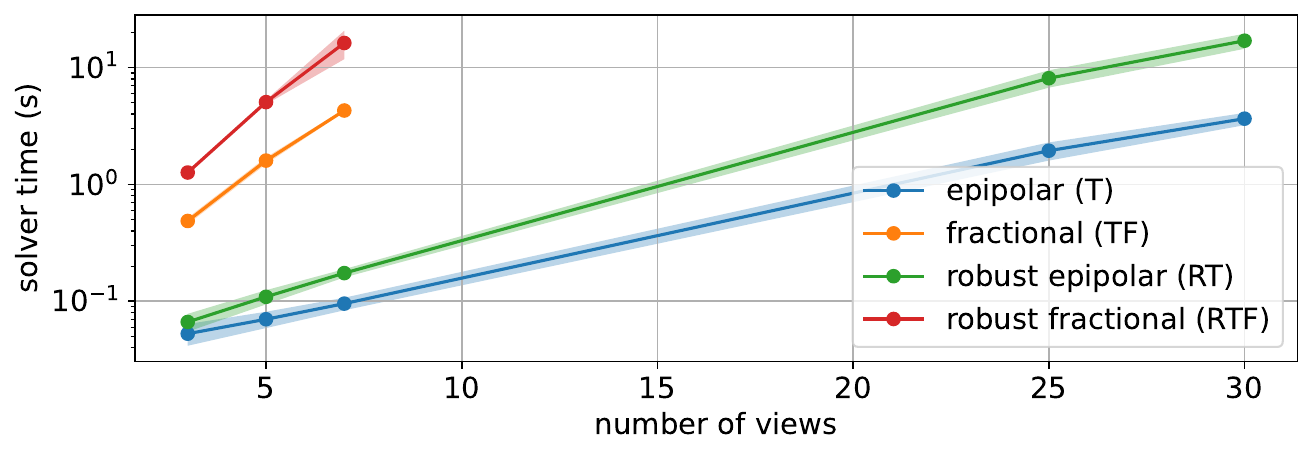} 
    \end{center}
    \vspace{-0.75cm}
    \caption{Average computation time for each solver, averaged over all noise levels and number of outliers.}
     \vspace{-0.65cm}
    \label{fig:solver-timings}
\end{figure}

\section{Experiments}
\label{sec:experiments}

We implement all relaxations using CVXPY \cite{diamond2016cvxpy} with the solver MOSEK \cite{mosek} using the setting $\texttt{MSK\_DPAR\_INTPNT\_CO\_TOL\_REL\_GAP} = 10^{-14}$,  all other solver parameters are left on their defaults. We find that working in units of pixels results in poorly conditioned solutions where $\hat z$ does not satisfy the constraints to high accuracy even for tight relaxations. To avoid this issue we use the change of variables $x_i \rightarrow \frac{1}{W}x_i$ and adjust the intrinsics accordingly. Since the scaling is the same for each point the optimal solution remains unchanged, but we get much closer to rank one solutions in practice due to the improved numerical stability.

\subsection{Simulated experiments}
\label{sec:synthetic}
We simulate triangulation problems as initially proposed in \cite{Hartley2009TriangulationSimulation} by placing $n$ cameras on a sphere of radius 2 and sample a point to be triangulated from the unit cube, see \cref{fig:triangulation-problems} for some examples. The same setup was also used for experiments in \cite{aholt2012qcqp, Cifuentes19}. For the reprojection model we simulate a pinhole camera with parameters from one of the cameras in the Reichstag dataset: width $W=2108$, height $H=1162$, focal length $f=1012.0027$ and principal point $p = (1054, 581)$. We simulate noisy observations by adding Gaussian noise with standard deviation $\sigma$ to the ground truth image coordinates. When generating an outlier we select a view at random and replace the measurement with a random point in the image.

We run the experiment for each method at various different noise levels and number of outliers. For each noise level we run \cref{eqn:robust-triangulation-relaxation} 750 times and \cref{eqn:robust-triangulation-fractional-relaxation} 120 times for $n=3, 5$ and 7 views and in each case add up to $n-2$ outliers. The percentage of tight relaxations and the estimation error can be seen in \cref{fig:experiments-error-tightness}. We also run \cref{eqn:robust-triangulation-relaxation} 45 times each for $n=25$ and 30 with 0, 10, 20 and 25 outliers, the results of which can be seen in \cref{fig:experiments-error-tightness-25}. We don't run \cref{eqn:robust-triangulation-fractional-relaxation} for these cases since we run into memory limitations with MOSEK. We set $c_i=200^2$ for all experiments.

From \cref{fig:experiments-error-tightness} we can see that in general the fractional relaxation in \cref{eqn:robust-triangulation-fractional} is significantly more stable than the epipolar relaxation \cref{eqn:robust-triangulation-relaxation}. In fact, across all experiments the fractional relaxation is tight in $99.92\%$ of cases. However, we can also note that the epipolar relaxation remains viable for lower noise levels, for instance in the case with $n=7$ views and 3 outliers the relaxations is tight in more than 90$\%$ of cases when the noise is below $\sigma\approx 40$px, after which the percentage of tight relaxations drop drastically.

As can be seen from the average solver timings in \cref{fig:solver-timings} the fractional relaxations is also over one order of magnitude slower than the epipolar relaxation, meaning that it is preferable to use \cref{eqn:robust-triangulation-relaxation} in cases where the quality of observations is known to be high, and fall back to \cref{eqn:robust-triangulation-fractional-relaxation} only if the epipolar relaxation is non-tight and spending extra time on reducing the error is desirable.

\subsection{Reichstag dataset}
\label{sec:reichstag-dataset}
We also validate our relaxations on the Reichstag dataset from \cite{Jin2020ImageMatching}. The dataset consits of 75 views of roughly 18k 3D points. We use the ground truth correspondences estimated by structure from motion as detailed in \cite{Jin2020ImageMatching} and generate each triangulation problem by selecting $n$ views which all observe a common point. We then add up to $n-2$ outliers by replacing the ground truth observations with a randomly selected keypoints in the same image. See \cref{fig:reichstag-triangulation} for an example point with $n=22$ views and 19 outliers.

\begin{figure}
    \begin{center}
        \vspace{-0.5cm}
        \includegraphics[width=\linewidth]{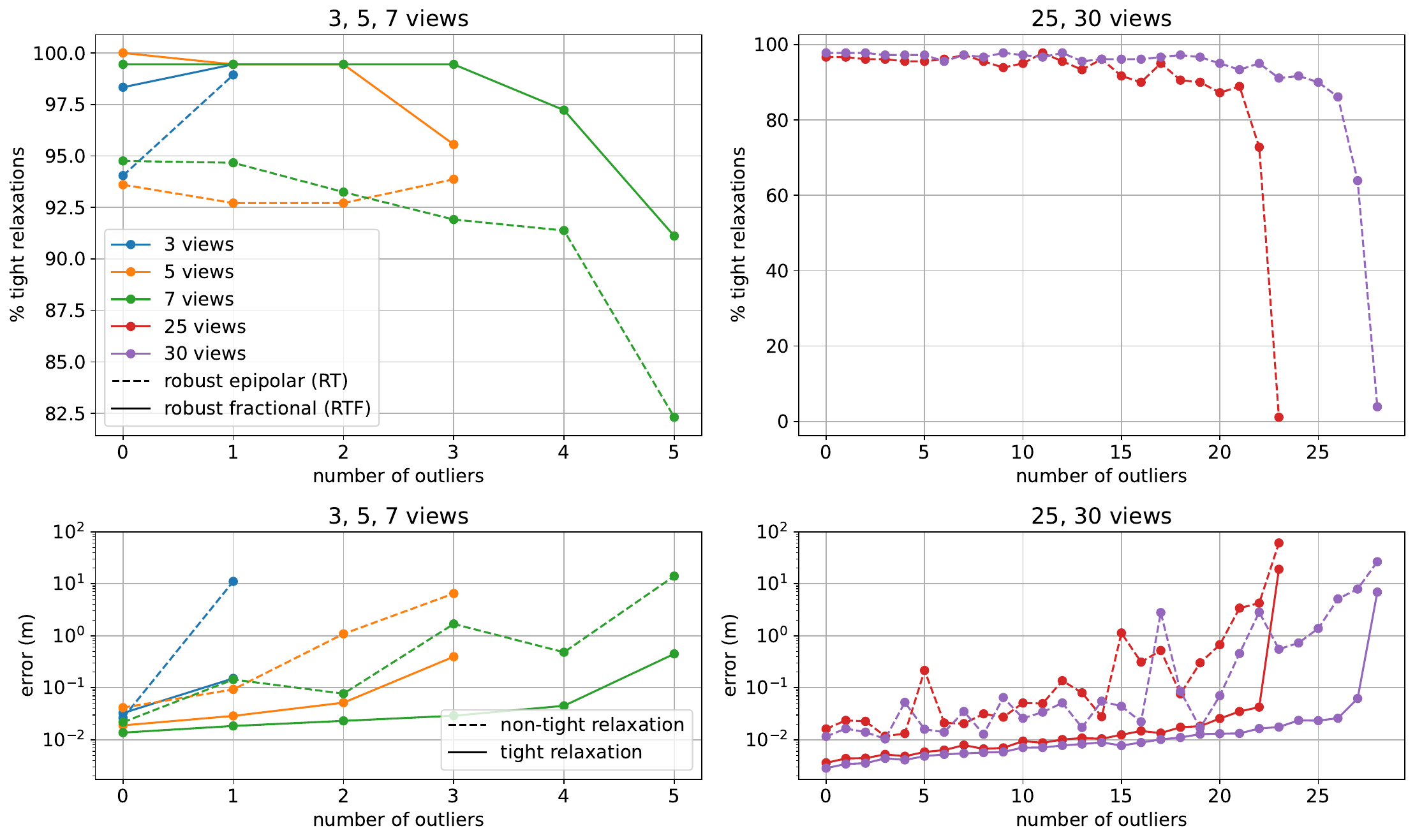} 
    \end{center}
    \vspace{-0.75cm}
    \caption{Average number of tight relaxation for \cref{eqn:robust-triangulation-relaxation} and \cref{eqn:robust-triangulation-fractional-relaxation} (top) and estimation error for tight and non-tight relaxations (bottom)  on the Reichstag dataset. See \cref{sec:reichstag-dataset} for details.}
    \label{fig:reichstag-experiments}
\end{figure}

For $n=3$, 5 and 7 views we run \cref{eqn:robust-triangulation-relaxation} 375 times and \cref{eqn:robust-triangulation-fractional-relaxation} 60 times for each possible number of outliers with $c_i = 10^2$. And similarly we run \cref{eqn:robust-triangulation-relaxation} 120 times for $n=25$ and 30 views. The results are summarized in \cref{fig:reichstag-experiments}.

Similarly to the simulated experiments we can note that the percentage of tight relaxations decreases steadily as more outliers are added, with a sharp drop when the number of inliers gets close to 2, with the fractional method outperforming the epipolar method.

\begin{figure}%
    \newcommand{\problemwidth}{\columnwidth}
    \centering
    \begin{subfigure}{\problemwidth}
        \includegraphics[width=\columnwidth]{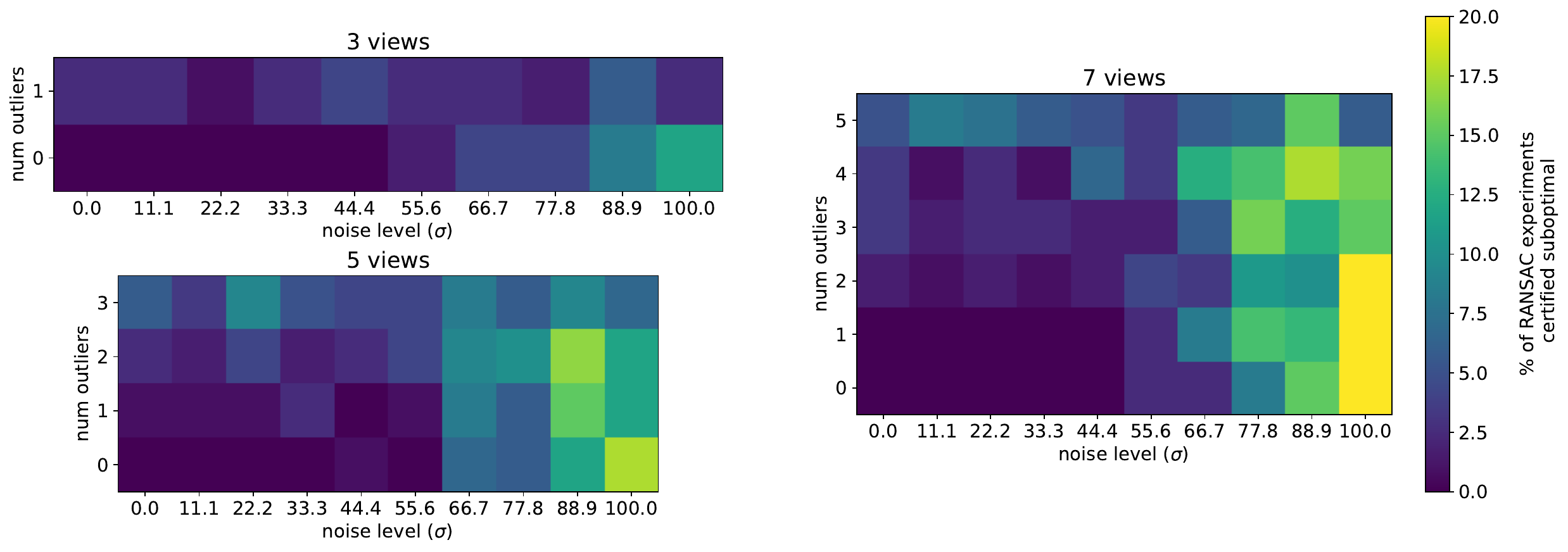}%
        \label{subfiga}%
    \end{subfigure}

    \begin{subfigure}{\problemwidth}
        \includegraphics[width=\columnwidth]{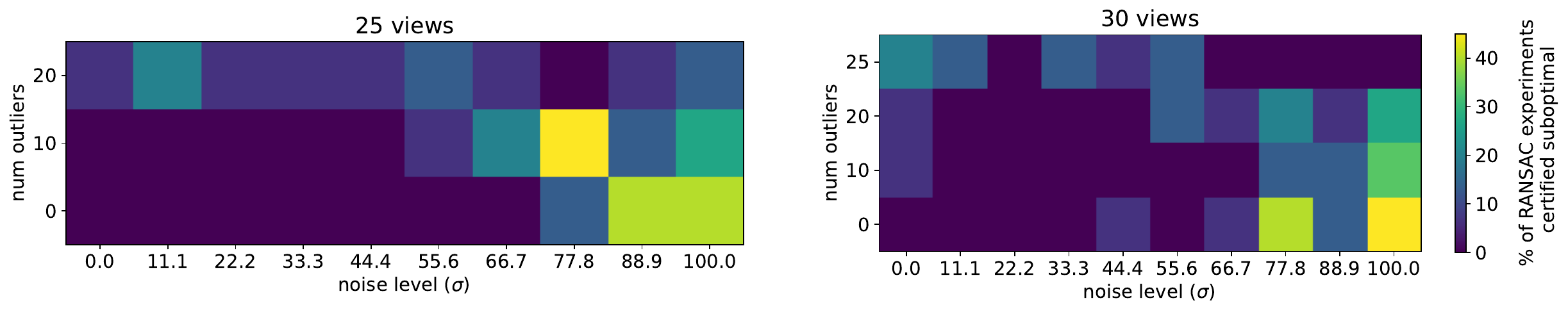}%
        \label{subfiga}%
    \end{subfigure}

    \caption{Color indicates the percentage of experiment trials where our robust relaxation (\cref{eqn:robust-triangulation-relaxation} for 3, 5 and 7 views and \cref{eqn:robust-triangulation-fractional-relaxation} for 25 and 30 views) found a lower cost solution than RANSAC. Details in \cref{sec:certifying-ransac}.}
     \vspace{-0.55cm}
    \label{fig:ransac-statistics}
\end{figure}

\subsection{RANSAC comparison}
\label{sec:certifying-ransac}
In many scenarios it is unrealistic to spend the computation time necessary for certifiable optimality at runtime, yet certifiably optimal algorithms can still provide valuable insight to algorithm developers in an offline fashion. In this section we will provide an illustrative how this principle can be applied to RANSAC-based triangulation. 

In particular, we compare our relaxations with exhaustive MLESAC \cite{MlesacTORR2000138} with TLS objective which we implement in the following way: 
first off, in order to eliminate the effect of randomness, we evaluate every possible pair of views. 
 For each pair of views we generate a candidate point by computing the optimal triangulation using \cref{eqn:triangulation-sdr}, which is always tight in the 2-view case (see \cite{aholt2012qcqp}), and evaluate the robust cost function in \cref{eqn:robust-triangulation-order2}. 
 From the candidate point achieving the minimal robust cost we further generate an additional candidate point by locally refining \cref{eqn:triangulation} on the inlier points.
 The final RANSAC estimate is then whichever of the candidate points obtains the lowest robust cost. 
 
 We run our RANSAC implementation on every triangulation problem from the simulated experiments in \cref{sec:synthetic} and compare against the results obtained by the certifiably optimal solvers. In cases where the relaxation is not tight we also refine the rounded solution on the inlier set. We do this for all simulated experiments and compare against our relaxations, see \cref{tab:ransac-comparison} for a breakdown of the number of cases where RANSAC was certified as optimal/suboptimal. In \cref{fig:ransac-statistics} we show the percentage of cases where the robust relaxation finds a better solution than RANSAC. For low noise and outlier levels RANSAC more or less always finds the globally optimal 
 solution, while for high noise levels there are many failure cases.

\begin{table}
    \resizebox{1\columnwidth}{!}{\begin{tabular}{cc|ccc}
        \midrule
        views & tight & relaxation better & same solution & RANSAC better
        \\ \hline
        3 & \cmark & 69 & 2331 & 0 \\
        3 & \xmark & 0 & 0 & 0 \\ \hline
        5 & \cmark & 257 & 4542 & 0 \\
        5 & \xmark & 1 & 0 & 0 \\ \hline
        7 & \cmark & 446 & 6744 & 0 \\
        7 & \xmark & 1 & 6 & 3 \\ \hline
        \bottomrule
        views & tight & relaxation better & same solution & RANSAC better
        \\ \hline
        25 & \cmark & 12 & 319 & 0 \\
        25 & \xmark & 32 & 58 & 29 \\ \hline
        30 & \cmark & 18 & 424 & 0 \\
        30 & \xmark & 31 & 84 & 43 \\ \hline
        \bottomrule
    \end{tabular}}
    \vspace{-0.3cm}
        \caption{Top: \cref{eqn:robust-triangulation-relaxation} vs. RANSAC for 3, 5 and 7 views. Bottom: \cref{eqn:robust-triangulation-fractional-relaxation} vs. RANSAC for 25 and 30 views.}
        \label{tab:ransac-comparison}
\end{table}

\section{Conclusion}
We proposed a global optimization framework for robust multiview triangulation.  To this end we derive semidefinite relaxations for triangulation losses that incorporate a truncated quadratic cost making them robust to both noise and outliers.  On synthetic and real data we confirm that provably optimal triangulations can be computed and relaxations remain empirically tight despite significant amounts of noise and outliers.

\subsection{Acknowledgments}
This work was supported by the ERC Advanced Grant SIMULACRON, by the Munich Center for Machine Learning and by the EPSRC Programme Grant VisualAI EP/T028572/1.

{\small
\bibliographystyle{ieee_fullname}
\bibliography{egbib}
}

\clearpage
\appendix

\section{Co-planar solution to epipolar relaxation}
\label{sec:coplanar}
\begin{figure}%
    \newcommand{\problemwidth}{.5\columnwidth}
    \centering
    \begin{subfigure}{\problemwidth}
        \centering
        \includegraphics[width=.9\columnwidth]{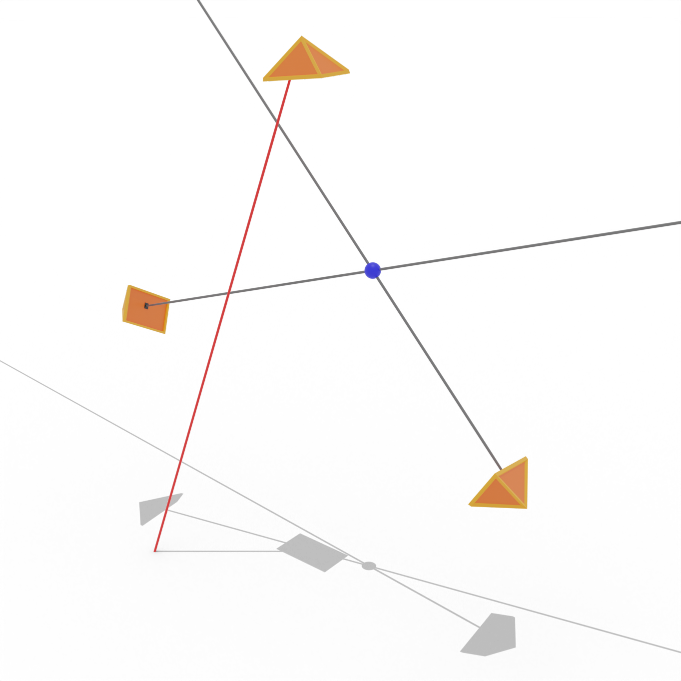}%
        \caption{Observations $\obs_i$}%
        \label{subfiga}%
    \end{subfigure}%
    \begin{subfigure}{\problemwidth}
        \centering
        \includegraphics[width=.9\columnwidth]{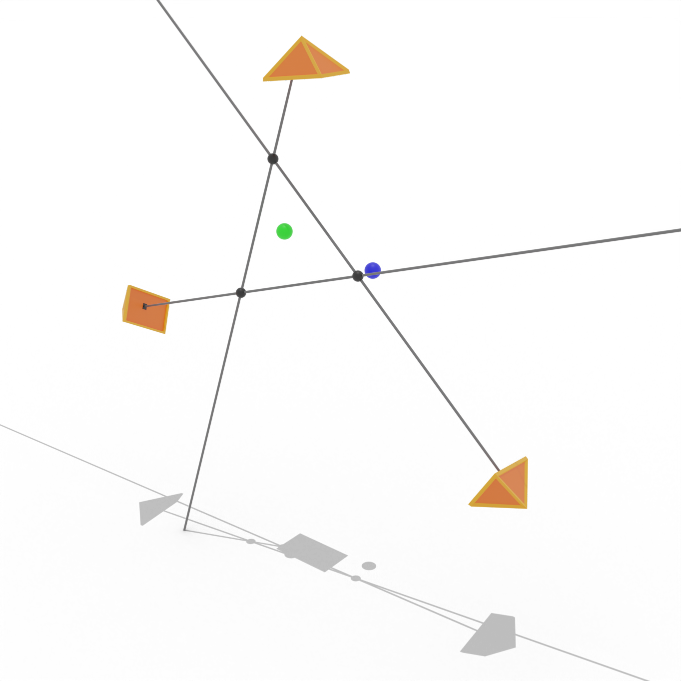}%
        \caption{Invalid estimated reprojections $\repr_i$}%
        \label{subfigb}%
    \end{subfigure}
    \caption{{\color{blue} Blue point}: ground truth. {\color{green} Green point}: invalid point estimate obtained by rounding the solution to \cref{eqn:robust-triangulation-relaxation}. Black points: intersection of viewing rays.}
    \label{fig:coplanar}
\end{figure}

\begin{figure}
    \begin{center}
        \includegraphics[width=\linewidth]{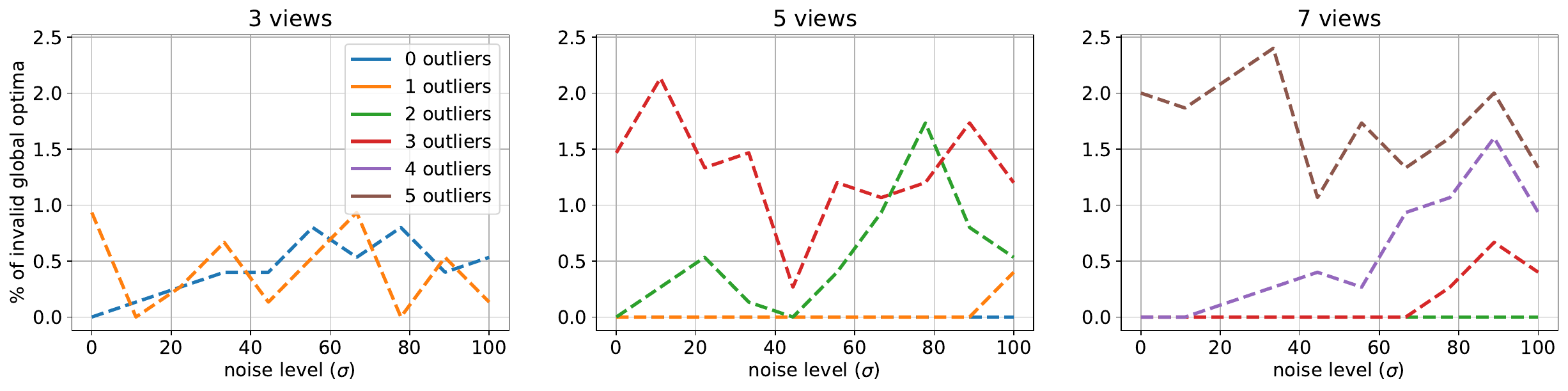} 
    \end{center}
    \vspace{-0.75cm}
    \caption{Percentage of invalid global optima found by \cref{eqn:robust-triangulation-relaxation} in the simulated experiments. We found none for 25 and 30 views.}
     \vspace{-0.65cm}
    \label{fig:coplanar-stats}
\end{figure}

When all camera centers are co-planar, any configuration of observations $x_i$ of viewing rays which lie in the camera plane will satisfy the epipolar constraints, despite not necessarily corresponding to the repeojection of a single 3D point. 
This means solutions to \cref{eqn:triangulation-sdr} and \cref{eqn:robust-triangulation-relaxation} might not correspond to valid 3D points, despite the relaxations being tight.
Following \cite{aholt2012qcqp}, we regard any solution with such invalid global optima as non-tight in all our experiments.

An example of a configuration where an invalid global optima occurs is shown in figure \cref{fig:coplanar}.
In the example the true 3D point lies close to the camera plane and there are two inlier observations with noise $\sigma = 0$ and one outlier whose viewing ray also lies close to the camera plane.
In this case adjusting the viewing rays such that they all lie in the camera plane produces a lower cost solution than correctly labelling the third view as an outlier. In contrast, \cref{eqn:robust-triangulation-fractional-relaxation} produces the correct solution, since the 3D point is explicitly parametrized.
In \cref{fig:coplanar-stats} we show the number invalid global optima found in the simulated experiments.

 See \cite{aholt2012qcqp} and \cite{astrom1997algebraic} for a more detailed discussion on when a solution of \cref{eqn:triangulation-poly} is guaranteed to generate a unique solution to \cref{eqn:triangulation}.

\section{Noise-free and outlier-free case}
\label{sec:theory}
In this section we will show that both our relaxations are tight in the noise-free and outlier-free case, we will also prove some of the criteria needed for local stability. 

We can verify whether a potential solution to \cref{eqn:qcqp} is globally optimal by computing the corresponding \textit{Lagrange multipliers}, as summarized in the following fact:
\begin{fact}
\label{fact:duality-gap}
If $\hat z\in\mathbb{R}^d$ satisfies the constraints of \cref{eqn:qcqp} (primal feasibility) and there are Lagrange multipliers $\hat\lambda\in\mathbb{R}$, $\hat\xi\in\mathbb{R}^k$ and a corresponding multiplier matrix $S(\hat\lambda,\hat\xi) = M + \sum_{i=1}^k\hat\xi_iA_i - \hat\lambda E$ satisfying:
\begin{enumerate}[i)]
    \item Dual feasibility: $S(\hat\lambda, \hat\xi) \succcurlyeq 0$
    \item Complementarity: $S(\hat\lambda, \hat\xi)\hat z = 0$
\end{enumerate}
then the relaxation \cref{eqn:primal} is tight and $\hat z$ is optimal for \cref{eqn:qcqp}.
\end{fact}

It might seem surprising that semidefinite relaxations of geometry problems in computer vision are empirically tight to such a large extent, but \cite{cifuentes2022local} provides some theoretical justification for this observation. They show for instance that under a smoothness condition  \cref{eqn:primal} will be a tight relaxation of \cref{eqn:qcqp} for problems that are close in parameter-space to solutions where the multiplier matrix has corank 1\footnote{\text{corank}(A) = n - \text{rank}(A) for an $n\times n$ matrix $A$.}. We will later show the corank 1 condition for the noise-free and outlier-free case of the triangulation problem, although we have not investigated the smoothness condition. We restate the main result in loose terms here:
\begin{fact}
\label{fact:local-stability}
If we, in addition to the conditions in \cref{fact:duality-gap}, have that $S(\lambda, \mu)$ is corank 1 and ACQ (which is a smoothness condition, see \cite{cifuentes2022local} Definition 3.1) holds, then the relaxation \cref{eqn:primal} is \textit{locally stable}, meaning that it will remain tight also for perturbed objective functions $M + \varepsilon \tilde M$ for small enough $\varepsilon$.
\end{fact}
The practical usefulness of \cref{fact:local-stability} comes from the consideration that it's often possible to show that the relaxation is tight and the stability conditions hold for noise-free measurements. This means that there is some surrounding region of noisy measurements for which the relaxation is tight as well. 

\subsection{Epipolar method}
\label{sec:local-stability-epipolar}
In the noise-free and outlier-free case we can show that the relaxation is tight with a corank 1 multiplier matrix:
\begin{theorem}
\label{thm:robust-epipolar}
The relaxation \cref{eqn:robust-triangulation-relaxation} is tight with a corank 1 multiplier matrix for noise-free and outlier-free measurements $\obs_i$, $i=1,\ldots,n$.
\end{theorem}
\begin{proof}
Partiton the lagrange multipliers as $\xi = (\varphi;\mu;\eta)$, where $\varphi_{ij}\in\mathbb{R}$ $\mu_i\in\mathbb{R}^2$ and $\eta\in\mathbb{R}$ corresponds to the constraints $(y_i\cat\theta_i)^TF_{ij}(y_j\cat\theta_j) = 0$, $\theta_iy_i=y_i$ and $\theta_i^2 = \theta_i$ respectively. Then we have:
\eq{
&S(\lambda, \varphi, \mu, \eta) =
\\ &F(\varphi) + \begin{pmatrix}
    I& -B(\obs_i - \mu_i) & -\mu
\\  * & \text{diag}(\|\obs_i\|^2 + 2\eta_i) & -\frac{1}{2}c - \eta
\\  * & * & \sum_{i=1}^nc_i - \lambda
\end{pmatrix}.
}
Where $F(\varphi) = \sum_{ij}\varphi_{ij}\bar F_{ij}$. Now let $\hat\lambda=\hat\varphi_{ij}=\hat\mu_i=0$ and $\hat\eta_i=\frac{1}{2}c_i$ to get:
\eq{
&\hat S = S(\hat\lambda, \hat\varphi, \hat\mu, \hat\eta) = S(0, 0, 0, \frac{1}{2}c) =
\\ &\begin{pmatrix}
    I & -B(\obs_i) & 0
\\  * & \text{diag}(\|\obs_i\|^2 + c_i) & -c
\\  * & * & \sum_{i=1}^nc_i
\end{pmatrix}.
}
This way, with $\hat z = (\tilde x; \mathbf{1}_n; 1)$ we have $\hat S\hat z = 0$. And furthermore, for arbitrary $x, \theta, \onevar$: 
\begin{equation*}
\begin{aligned}
    &(x\cat\theta\cat \onevar)^T\hat S(x\cat\theta\cat \onevar)=
    \\&= \sum_{i=0}^n\bigg(\|x_i\|^2 - 2\theta_i\obs_i + \theta_i^2(\|\obs_i\|^2\!+\!c_i) - 2c_i\theta_i\onevar + c_i\onevar^2\bigg)
    \\&= \sum_{i=0}^n\bigg(\|x_i - \theta_i\obs_i\|^2 + c_i(\onevar - \theta_i)^2\bigg)\geq 0
\end{aligned}
\end{equation*}
so $\hat S$ is positive semidefinite. So the relaxation is tight by \cref{fact:duality-gap}. And since the only nonzero solution to $(x\cat\theta\cat\onevar)^T\hat S(x\cat\theta\cat\onevar) = 0$ up to scale is $(x\cat\theta\cat\onevar) = \hat z$ we have that $\hat S$ is corank 1.
\end{proof}

We emphasize that since we don't have a proof for the ACQ condition we haven't fully proved local stability. But we include the partial results in case they are useful for future works.

\subsection{Fractional method}
\label{sec:local-stability-fractional}
In this section we will prove two of the criteria required for local stability for the robust fractional method \cref{eqn:robust-triangulation-fractional-relaxation} for noise-free and outlier-free measurements. Local stability for the non-robust case was shown already in \cite{Cifuentes19} but we will provide an alternate proof here in our notation, since it will lead into the extension to the robust case. For this we will need the stronger version of \cref{fact:local-stability}, which we will restate here loosely (see \cite{cifuentes2022local} Theorem 4.5 for more details). Using the definition $A(\xi) = \sum_{i=1}^k\xi_iA_i$:
\begin{fact}
    If we, in addition to the conditions in \cref{fact:duality-gap}, have that:
    \begin{enumerate}[(i)]
        \item (ACQ) ACQ holds
        \item (smoothness) The the constraint set is smooth with respect to perturbations to the constraints
        \item (non-branch point) The nullspace of the multiplier matrix and the tangent space of the constraint-set at the optimum don't intersect nontrivially: $\textrm{\emph{ker}}(\hat S)\cap T_{\hat z} = \{0\}$
        \item (restricted slater) There exists $\xi'$, $\lambda'$ such that $A(\xi')-\lambda'E$ is positive definite on the subspace of vectors $z_\perp$ for which $\hat Sz_\perp = 0$ and $\hat z^Tz_\perp \neq 0$. In other words the part of the nullspace of $\hat S$ which is orthogonal to the solution $\hat z$.
    \end{enumerate}
\end{fact}

The tangent space in (iii) is given by $T_{\hat z} = \textrm{ker}(\hat z^TA_1\cat\ldots\cat \hat z^TA_k\cat \hat z^TE)$. 

\subsection{Non-robust version}
We will show (iii-iv) for a version of \cref{eqn:triangulation-fractional-relaxation} with somewhat less constraints, noting that if we show (iii-iv) for the problem with less constraints we can then add in the remaining constraints back in and set the corresponding multipliers to zero to show that (iii-iv) holds for the original problem as well. Note however again that since we don't show (i-ii) the full proof is incomplete and is left for future work.

\begin{theorem}
\label{thm:fractional}
The fractional relaxation \cref{eqn:triangulation-fractional-relaxation} is tight and (iii-iv) hold for noise-free and outlier-free measurements $\obs_i$, $i=1,\ldots,n$.
\end{theorem}
\begin{proof}
We start by partitioning the Lagrange multipliers as $\xi = (\varphi\cat \alpha)$. Where $\varphi = (\varphi_1\cat\ldots\cat \varphi_{2n})$, and each $\varphi_i\in\mathbb{R}^4$ contains the multipliers corresponding to $i$th reprojection constraint multiplied by the entries of $\bar X$ (recall that there are two reprojection constraints per observation). Note that in the original formulation we also multiply by all the entries of $\repr\otimes\hompoint$ as well, but as we will see these are not necessary for the proof to hold. And $\alpha$ corresponds to the Kronecker product constraints.

Since the observations $\obs$ are noise free we can denote the corresponding unique\footnote{assuming the observations are not degenerate, \eg not all on a line.} 3D point in homogeneous coordinates as $\hat\point\in\mathbb{R}^4$, normalized such that $\|\hat\point\|=1$. It will be convenient to introduce the reparametrization $u=\obs$ which is the same as the observation vector, except partitioned such that $u=(u_1\cat\ldots\cat u_{2n})$, $u_i\in\mathbb{R}$, \ie $u_{2i + k} = \obs_{ik}$ for $i=1,\ldots,n$, $k=1,2$. The primal optimum is then obtained at $\hat z = \bar u\otimes\hat\point$, which is verified by setting $\hat\xi=\hat\lambda=0$ to get $\hat S\hat z = (M_{\obs}\otimes I_4)(\bar u\otimes\hat\point) = (M_{\obs}\bar u)\otimes\hat\point = 0$.

We then note that, due to the properties of the Kronecker product\footnote{For matrices $A\in\psdset{n}, B\in\psdset{m}$ with eigenvalues $\alpha_i$, $\beta_j$ the eigenvalues of the Kronecker product $A\otimes B$ are given by the products of the eigenvalues $\alpha_i\beta_j$ for $i=1,\ldots,n$, $j=1,\ldots,m$.} and that $M_{\obs}$ is positive semidefintie with corank 1, we have that $\hat \psd = M_{\obs}\otimes I_4$ is positive semidefinite with corank 4. So the conditions of \cref{fact:duality-gap} are satisfied and the relaxation is tight.

Since the nullspace $\textrm{ker}(\hat S)$ is 4-dimensional and contains the four orthogonal vectors $\hat z = \bar u\otimes\hat\point$ and $\hat z_l = \bar u\otimes\hat\point_l$ where $\hat\point^T\hat\point_l = 0$, $\hat\point^T_l\hat\point_k=0$ for $k\neq l=1,2,3$ we can parametrize $z_\perp$ from (iv) as $z_\perp = \bar u\otimes\hat\point_\perp$ where $\hat\point_\perp^T\hat\point = 0$.

For (iii) we need to show that the vectors that span $\textrm{ker}(\hat S)$ are not in $T_{\hat z}$, \ie for any $z\in\textrm{ker}(\hat S)$ either that $\hat z^T A_iz \neq 0$ for some constraint $i$, or that $\hat z^TEz\neq 0$. This is the case since $\hat z^TE\hat z = 1 \neq 0$ and, letting $K_{ijst}$ be the Kronecker constraint matrix corresponding to index $st$ of block $ij$, $\hat z^TK_{ijst}z_l = u_iu_j(\hat\point_s\hat\point_{lt} - \hat\point_t\hat\point_{ls})$ is nonzero for at least some index $ijst$ unless $u=0$ or $\hat\point$ and $\hat\point_l$ are parallel, which is not the case by construction.

To show (iv), we set $\alpha'=\lambda'= 0$ and $\varphi_i' = u_i b_i - a_i$, and verify that with $z_\perp$ as above:
\eq{
\label{eqn:mu-pd}
    z_\perp^T A(\varphi', 0)z_\perp &= \sum_{i=1}^{2n}\hat\point_\perp^T\varphi_i'(u_ib_i - a_i)\hat\point_\perp
    \\              &= \sum_{i=1}^{2n}((u_ib_i - a_i)^T\hat\point_\perp)^2>0
}
where the final strict inequality follows from the fact that each term is strictly positive as $(u_ib_i - a_i)^T\hat\point=0$ by the original constraints and $\hat\point_\perp$ is orthogonal to $\hat\point$.
\end{proof}

We note that, while not all constraints used in \cref{eqn:triangulation-fractional-relaxation} are required for (iii-iv) to hold, we have found some cases where adding the additional constraints results in a tighter relaxation in the presence of noise, so we used the full set of constraints in our experiments.

\subsection{Robust version}
We now move on to the robust fractional method
\begin{theorem}
\label{thm:robust-fractional}
The fractional relaxation \cref{eqn:robust-triangulation-fractional-relaxation} is tight and (iii-iv) hold for noise-free and outlier-free measurements $\obs_i$, $i=1,\ldots,n$.
\end{theorem}
\begin{proof}
Partition the Lagrange multipliers as $\xi = (\varphi\cat\mu\cat\eta\cat\alpha)$, where as in \cref{thm:fractional} $\varphi$ corresponds to the reprojection constraints and $\alpha$ corresponds to the Kronecker constraints. We let $\mu\in\mathbb{R}^{32n}$ correspond to the constraints $\hompoint_s\hompoint_t(y_{ik}\theta_i - y_{ik})= 0$ for $s,t=1,2,3,4$, $k=1,2$ and $i=1,\ldots,n$. And finally we similarly have that $\eta\in\mathbb{R}^{16n} = (\eta_1\cat\ldots\cat\eta_n)$, $\eta_i\in\mathbb{R}^{16}$ corresponds to the constraints $\hompoint_s\hompoint_t(\theta_i^2 - \theta_i) = 0$. For each view $i$ we collect the corresponding subset of $\eta$ into a $4\times4$ matrix $H_i$ defined such that $\hompoint^TH_i\hompoint = \sum_{s,t=1}^4\eta_{ist}\hompoint_s\hompoint_t$.

To verify the global optimum we start by setting $\hat z = \bar u_\theta\otimes\hat\point$ where $u_\theta = (\obs\cat 1_n)$. We then note that the constraint matrices for for the $\eta_i$-constraints can be written as a Kronecker product to get:
\eq{
    S(0, 0, \eta, 0) = M_{\obs}^c\otimes I_4 + \sum_{i=1}^nT_i\otimes H_i
}
where each $T_i\in\psdset{3n+1}$ is defined such that $\bar y_\theta^TT_i\bar y_\theta = \theta_i^2 - \theta_i$ for arbitrary $y_\theta$ as in \cref{sec:robust-fractional}. We then set $\hat\eta$ such that $\hat H_i = c_iI_4$ and $\hat\varphi=\hat\mu=\hat\alpha=\hat\lambda=0$ to get:
\eq{
    \hat S = S(0, 0, \hat\eta, 0) = (M_{\obs}^c + \sum_{i=1}^nc_iT_i)\otimes I_4.
}
Now, by the same argument as in \cref{thm:robust-epipolar} the matrix $M_{\obs}^c + \sum_{i=1}^nc_iT_i$ is positive semidefinite with corank 1, so $\hat S$ is positive semidefinite with corank 4. Meaning that the conditions of \cref{fact:duality-gap} are satisfied. (iii) also follows using the same argument based on the Kronecker constraints as in \cref{thm:fractional}.

Finally, for (iv) we note that $\textrm{ker}(\hat S)$ is spanned by $\hat z$ and $\hat z_l = \bar u_\theta\otimes\hat\point_l$, $l=1,2,3$, so by setting $\mu'=\eta'=\alpha'=\lambda'=0$ and $\varphi'_i = u_ib_i - a_i$ restricted slater for $\hat S$ follows in the same way as in \cref{eqn:mu-pd}.

\end{proof}

\end{document}


\title{Appendix - Semidefinite Relaxations for Robust Multiview Triangulation}

\author{First Author\\
Institution1\\
Institution1 address\\
{\tt\small firstauthor@i1.org}
\and
Second Author\\
Institution2\\
First line of institution2 address\\
{\tt\small secondauthor@i2.org}
}
\maketitle
\appendix
\section{Local stability of }